%% file: main.tex
\documentclass[10pt,twocolumn,letterpaper]{article}

\usepackage{iccv}              

\input{preamble}

\definecolor{iccvblue}{rgb}{0.21,0.49,0.74}
\usepackage[pagebackref,breaklinks,colorlinks,allcolors=iccvblue]{hyperref}

\usepackage{multirow}
\usepackage{graphicx}
\usepackage{amsmath}
\usepackage{utfsym}
\usepackage{siunitx}    
\usepackage{amssymb} 
\usepackage{pifont}  
\usepackage{listings}
\usepackage{array} 

\def\paperID{762} 
\def\confName{ICCV}
\def\confYear{2025}

\title{\textcolor{iccvblue}{TRCE}:
\textcolor{iccvblue}{T}owards
\textcolor{iccvblue}{R}eliable Malicious
\textcolor{iccvblue}{C}oncept
\textcolor{iccvblue}{E}rasure \\
in Text-to-Image Diffusion Models}

\author{Ruidong Chen$^1$,~
        Honglin Guo$^1$,~
        Lanjun Wang$^{2\ast}$,~
        Chenyu Zhang$^2$,~
        Weizhi Nie$^1$,~
        An-An Liu$^{1\ast}$ \\
        $^1$The School of Electrical and Information Engineering, Tianjin University\\
        $^2$The School of New Media and Communication, Tianjin University \\
}

\begin{document}
\maketitle

\renewcommand{\thefootnote}{}
\footnotetext{$^*$ Corresponding author: Lanjun Wang\{wang.lanjun\}@outlook.com,\\~An-An Liu\{anan0422\}@gmail.com}

\setlength{\stripsep}{-26pt}

\begin{strip}
    \centering
    \vspace{-8mm}
    \includegraphics[width=1\textwidth]{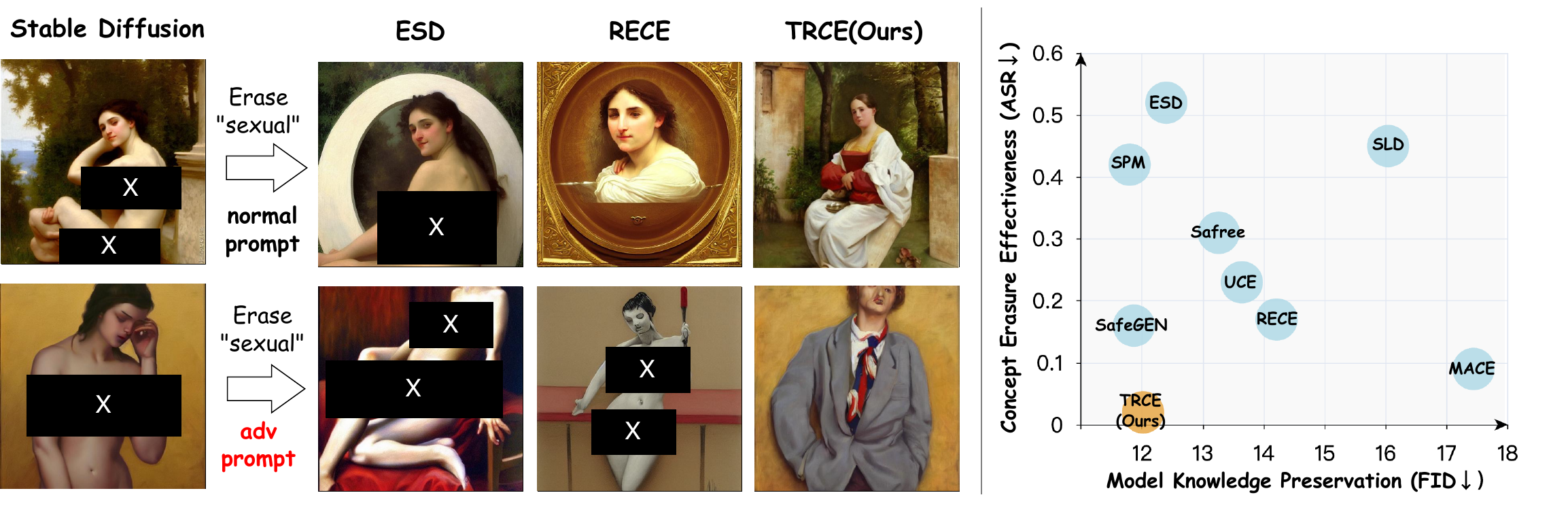}
    \label{fig1}
    \vspace{-10mm}
    \captionof{figure}{We propose TRCE to achieve reliable malicious concept erasure in text-to-image diffusion models. 
    Left~) TRCE effectively erases malicious concepts (e.g., “sexual”) 
    , even under adversarial prompts~\cite{p4d,ring-a-bell,unlearndiff,mma}. 
    Right~) 
    Compared to previous methods~\cite{esd,uce,rece,spm,sld, mace,yoon2024safree,li2024safegen}, TRCE reaches an outperform trade-off between erasure effectiveness and knowledge preservation, effectively removing malicious concepts while minimizing the impact on the irrelevant content. }
\vspace{12mm}
\end{strip}

\input{sec/0_abstract}    
\input{sec/1_intro}
\input{sec/2_related}
\input{sec/3_pre}
\input{sec/4_method}

\input{sec/5_exp}

\input{sec/6_conclusion}

{
    \small
    \bibliographystyle{ieeenat_fullname}
    \bibliography{main}
}
\appendix
\input{X_suppl}

\end{document}

%% file: preamble.tex
\usepackage{cuted}
\usepackage{adjustbox}
\usepackage{bm}

\newcommand{\red}[1]{{\color{red}#1}}


%% file: sec/0_abstract.tex
\begin{abstract}
Recent advances in text-to-image diffusion models enable photorealistic image generation, but they also risk producing malicious content, such as NSFW images. To mitigate risk, concept erasure methods are studied to facilitate the model to unlearn specific concepts. However, current studies struggle to fully erase malicious concepts implicitly embedded in prompts (e.g., metaphorical expressions or adversarial prompts) while preserving the model's normal generation capability. To address this challenge, our study proposes TRCE, using a two-stage concept erasure strategy to achieve an effective trade-off between reliable erasure and knowledge preservation. Firstly, TRCE starts by erasing the malicious semantics implicitly embedded in textual prompts. By identifying a critical mapping objective(i.e., the [EoT] embedding), we optimize the cross-attention layers to map malicious prompts to contextually similar prompts but with safe concepts. This step prevents the model from being overly influenced by malicious semantics during the denoising process. Following this, considering the deterministic properties of the sampling trajectory of the diffusion model, TRCE further steers the early denoising prediction toward the safe direction and away from the unsafe one through contrastive learning, thus further avoiding the generation of malicious content. Finally, we conduct comprehensive evaluations of TRCE on multiple malicious concept erasure benchmarks, and the results demonstrate its effectiveness in erasing malicious concepts while better preserving the model's original generation ability. The code is available at: \href{https://github.com/ddgoodgood/TRCE}{http://github.com/ddgoodgood/TRCE}.

\noindent \red{\textbf{CAUTION: This paper includes model-generated content that may contain offensive material.}}
\end{abstract}

%% file: sec/1_intro.tex
\section{Introduction}
Recently, large-scale text-to-image~(T2I) diffusion models~(DMs)~\cite{ldm,sdxl,sd3,imagen} have attracted widespread attention due to their capabilities to generate highly realistic images. 
Trained on extensive internet-sourced data, these models acquire the ability to produce diverse visual concepts from textual prompts.
However, due to the existence of toxic training data, these models also learn to generate inappropriate content. 
Consequently, they may be misused to produce NSFW (Not Safe For Work) images with prompts containing inappropriate concepts.
To address these safety concerns, researchers have proposed various mechanisms for enhancing safety~\cite{zhang2025adversarial}, such as filtering out toxic data and retraining the models~\cite{sd21}, employing safety checkers to filter outputs~\cite{red-safety}, and applying safeguard strategies to guide the generation process~\cite{sld,yoon2024safree}.
To further enhance safety, \textit{concept erasure~(CE)}~\cite{esd, ca,fmn,spm,li2024safegen,rece,uce, mace} has been proposed to limit the ability of DMs to generate specific concepts.
For example, if the concept ``sexual" were erased by CE, when the prompt includes keywords such as ``nudity, erotic, porn, etc.", the model will generate images of clothed people instead.

However, due to insufficient erasure, current CE methods are still not effective enough in preventing the generation of NSFW content~\cite{ring-a-bell,unlearndiff,p4d,mma}. 
Existing methods are typically designed to eliminate specific keywords by fine-tuning~\cite{esd,ca,fmn,spm,mace,li2024safegen} or editing~\cite{uce,rece,mace} model parameters. 
As a result, they struggle to erase malicious concepts, as they are often expressed metaphorically or associatively, without using keywords~(e.g., the prompt in the second row of Fig.~\ref{fig2}). 
In most scenarios, the semantics of malicious concepts are implicitly embedded in the prompts, which poses challenges for erasing malicious concepts.
To eliminate such implicit semantics, existing methods often degrade the model's original generation ability on irrelevant content to achieve stronger erasure reliability~(Fig.~1 Right).
Overall, the inability to \textbf{balance reliable concept erasure with knowledge preservation} constitutes the main limitation that prevents current CE methods from being effectively applied to eliminate malicious concepts.

To address these limitations, this study proposes \textbf{TRCE}, which is based on a two-stage erasure design, aiming to ensure both reliable erasure and knowledge preservation through the cooperation of the two stages.
Firstly, the TRCE begins with the ``\textit{Textual Semantic Erasure}"~(Sec.~\ref{method:stage1}), mitigating the influence of malicious semantics from the input prompt on the generation process.
Recognizing the influence of special embeddings~\cite{clip,devlin2018bert} on generation~(as illustrated in Fig.~\ref{fig2}), we note that the [EoT] (End of Text) embeddings capture semantics from the entire prompt and significantly contribute to the generation of salient regions~\cite{layoutfree}.
By leveraging this insight, we can move away from existing methods~\cite{uce,rece,mace} that directly map keyword embeddings to unrelated ones.
Instead, we select [EoT] as the mapping objective and employ a closed-form solution~\cite{uce,rece} to modify the cross-attention matrices. 
Mapping prompts with malicious concepts to contextually similar prompts but with safe concepts, thus achieving the effective erasure of implicitly embedded malicious semantics. 

On the basis of the first stage, TRCE further proposes ``\textit{Denoising Trajectory Steering}"~(Sec.~\ref{method:stage2}) to avoid the final generation of malicious visual content.
Typically, during the denoising process, a diffusion model first forms outlines of prominent regions and then produces specific visual details at the turning point~\cite{ssb}. 
Based on this behavior, at an early sampling stage of generation, we propose using a reference model to provide both safe and unsafe predictions before the turning point.
Specifically, a contrastive loss is employed to steer the denoising prediction toward the safe direction and away from the unsafe one.
As shown in Fig.~\ref{fig:method2}, take advantage of the deterministic properties~\cite{ddim} of diffusion model's sampling trajectory, 
since the malicious semantics of generation guidance have been mostly eliminated in the first stage, by just adjusting the model's early denoising, the model's sampling trajectory can deviate from unsafe visual patterns without significantly affecting the image content, thus achieving better knowledge preservation ability for the concept erasure.

We summarize the contribution of TRCE as follows:
\begin{figure}[t]
    \centering
\includegraphics[width=1\linewidth]{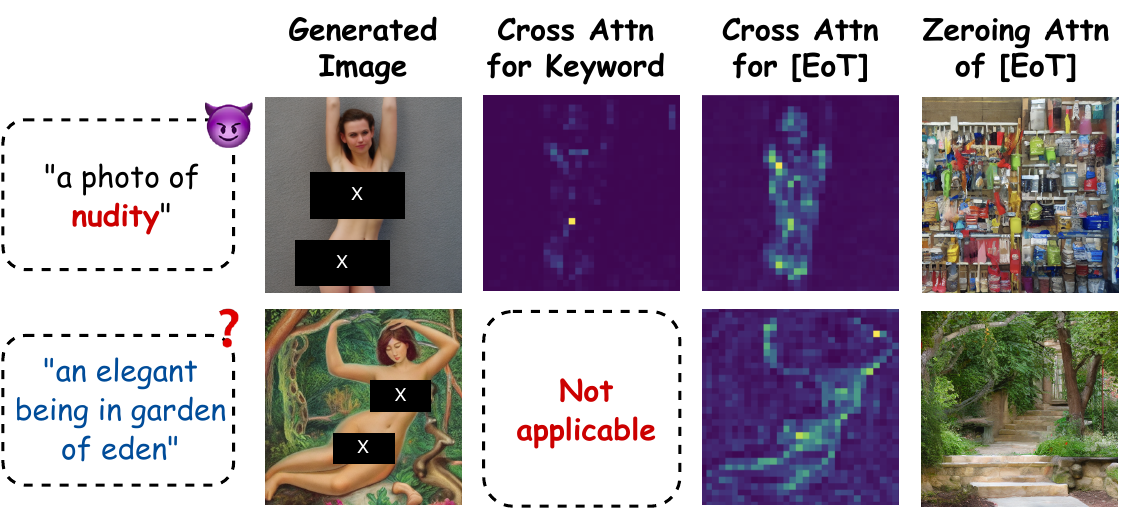}
    \caption{
    Due to the attention mechanism, special embeddings~\cite{clip,devlin2018bert} of input prompts~(e.g. [EoT]) carry rich semantics about concepts and pay attention to the semantics of salient regions~\cite{eotget}. 
    By eliminating the influence of the [EoT] token on generation (i.e., setting the attention map corresponding to [EoT] to 0), the image content will be dramatically affected. 
    }
    \label{fig2}
    \vspace{-5mm}
\end{figure}
\begin{itemize} 
    \item TRCE proposes a collaborated two-stage erasure strategy, achieving reliable malicious concept erasure while taking minimalist effects on the model's generation ability.
    \item TRCE identifies a critical mapping objective for textual semantic erasure, specifically the [EoT] embeddings, which achieves effective erasure of the implicitly embedded malicious semantics.
   \item TRCE proposes an effective denoising trajectory steering strategy, which optimizes the model's early denoising through contrastive fine-tuning.
    \item We conduct comprehensive evaluations of TRCE across various malicious content generation scenarios, including network prompts~\cite{sld}, adversarial prompts~\cite{unlearndiff,ring-a-bell,mma,p4d}, and multi-concept erasure~\cite{sld}, along with ablation studies. The results demonstrate the effectiveness of TRCE in conducting reliable malicious concept erasure.

\end{itemize}

%% file: sec/2_related.tex
\section{Related Work}
\subsection{Concept Erasure}
To prevent diffusion models from being misused for generating unsafe content, recent research has begun exploring methods to remove unsafe concepts from the model.
These methods can be categorized into two types: inference-time guidance~\cite{sld,yoon2024safree}, and concept erasure~\cite{ca,esd,fmn,spm,mace,uce,rece,li2024safegen}. 
Compared to inference-time guidance, concept erasure typically involves optimizing model parameters to mitigate the model’s ability to generate specific concepts, providing better performance in preventing unsafe content.
In early research related to this task, ESD~\cite{esd} and CA~\cite{ca} first proposed using the model's knowledge to erase their concepts, i.e., fine-tuning the model with specific concept prompts to align its noise predictions with those of unrelated concepts. 
To improve erasure efficiency, SPM~\cite{spm} and MACE~\cite{mace} suggest fine-tuning only a small adapter~\cite{hu2021lora} to perform concept erasure. 
Considering the critical role of cross-attention~\cite{attention,transformer} in concept generation within denoising networks~\cite{ddpm}. UCE~\cite{uce} first introduced a closed-form solution to directly optimize the attention matrix of the cross-attention layers, mapping concept keywords to unrelated concepts, which achieves fast erasure with only seconds per concept. 
Building upon UCE, RECE~\cite{rece} introduces an iterative erasure process based on UCE~\cite{uce} to enhance the robustness of concept erasure, which is the first study to discuss enhancing the robustness of concept erasure facing adversarial prompts~\cite{ring-a-bell,p4d,unlearndiff,mma}, however, it only improves the erasure robustness to a limited extent. 
In this work, we propose TRCE, which aims to further improve the reliable of concept erasure while better preserving the model's existing knowledge.

\subsection{Red-Teaming Tools}
With the rapid development of generative models, recent studies propose red-teaming tools~\cite{atk1_gcma,atk2,atk3,atk4} for discovering security vulnerabilities of models. 
Based on whether the attack accesses the internal mechanisms of the diffusion model, existing attack methods can be categorized into model-aware~\cite{p4d,unlearndiff} and model-agnostic~\cite{mma,ring-a-bell} approaches. 
Representative model-aware attack methods include P4D~\cite{p4d} and UnlearnDiff~\cite{unlearndiff}. These approaches typically target a malicious image by leveraging an alignment loss to optimize multiple token embeddings, which are subsequently combined into an adversarial prompt.
In contrast, MMA~\cite{mma} and Ring-a-Bell~\cite{ring-a-bell} are model-agnostic methods. Instead of directly interacting with T2I models, they rely solely on an open-source CLIP~\cite{clip} text encoder to optimize adversarial prompts by an alignment loss with malicious prompts within the CLIP feature space.
By mining the relevant adversarial prompts that lead to the generation of inappropriate content, these methods can identify vulnerabilities overlooked by concept erasure models, thereby guiding these models to regenerate malicious content. 
To address this challenge, TRCE aims to further enhance the reliability of the concept erasure, ensuring sufficient robustness even with adversarial prompts.

%% file: sec/3_pre.tex
\section{Preliminaries}
This work focuses on erasing concepts in latent diffusion models (LDMs)~\cite{ldm,sd14}, which have emerged as powerful tools for generating high-quality images. In this section, we introduce the preliminaries of LDMs, and relevant components we used to conduct erasure.

\noindent \textbf{Cross-attention Layers}: 
In LDMs, an image $x_0$ is first mapped into a latent representation $z_0$ using an encoder $\mathcal{E}$. 
The model then utilizes a denoising U-Net~\cite{unet}, denoted as $\epsilon_{\theta}$, to learn the denoising process~\cite{ddpm} in the latent space. 
To generate specific visual concepts, a pivotal aspect of LDMs is their ability to generate images conditioned on textual information. 
This is achieved through a CLIP~\cite{clip} text encoder $\tau_{\theta}$, which transforms a text prompt $y$ into the embedding
$\bm{e} = \tau_{\theta}(y) = \{e^{SoT},e^p_0,...,e^p_{l-1},e^{EoT}_0,...,e^{EoT}_{N-l-2}\}$,
where SoT and EoT are special tokens of start and end of the text~\cite{clip,devlin2018bert} respectively, $N$ is the maximum size processed by $\tau_{\theta}$, and $l$ is the size of words in the prompt. $\bm{e}$ is then incorporated into the generation via a cross-attention mechanism:
\begin{equation} 
\label{eq} 
\operatorname{Attn}(Q, K, V) = \operatorname{Softmax}\left(\frac{Q K^{T}}{\sqrt{d}}\right) V, 
\end{equation} 
where $Q = W_{Q} \cdot \varphi(z_t)$, $K = W_{K} \cdot \bm{e}$, $V = W_{V} \cdot \bm{e}$, and $\varphi(z_t)$ represent the hidden states in the U-Net at the diffusion step $t$ with the latent variable $z_t$.

\noindent \textbf{Classifier-free Guidance}:
To enhance the fidelity and controllability of generated images, LDMs employ the classifier-free guidance~(c.f.g) mechanism~\cite{class-free,glide}. 
This approach involves training the model to predict noise in both conditioned (on the text prompt $y$) and unconditioned (on the null text $\emptyset$) scenarios, and the training objective becomes:
\begin{equation}
\mathcal{L}_{\text{c.f.g.}} = 
\begin{cases} 
 \mathbb{E}_{z_0, y, \epsilon, t}  \left[ \left| \epsilon_{\theta} \left( z_t, y, t \right) - \epsilon \right|^2 \right], & \text{with } (1-p), \\
 \mathbb{E}_{z_0, y, \epsilon, t} \left[ \left| \epsilon_{\theta} \left( z_t, \emptyset, t \right) - \epsilon \right|^2 \right] , & \text{with } p.
\end{cases}
\end{equation}
where $z_t$ represents the latent feature at the time step $t$ obtained from $z_0$ by applying the noise $\epsilon$. The model is trained with probability $p$ to discard condition $y$, but trained unconditionally.

\begin{figure*}[t]
    \centering
    \includegraphics[width=1\textwidth]{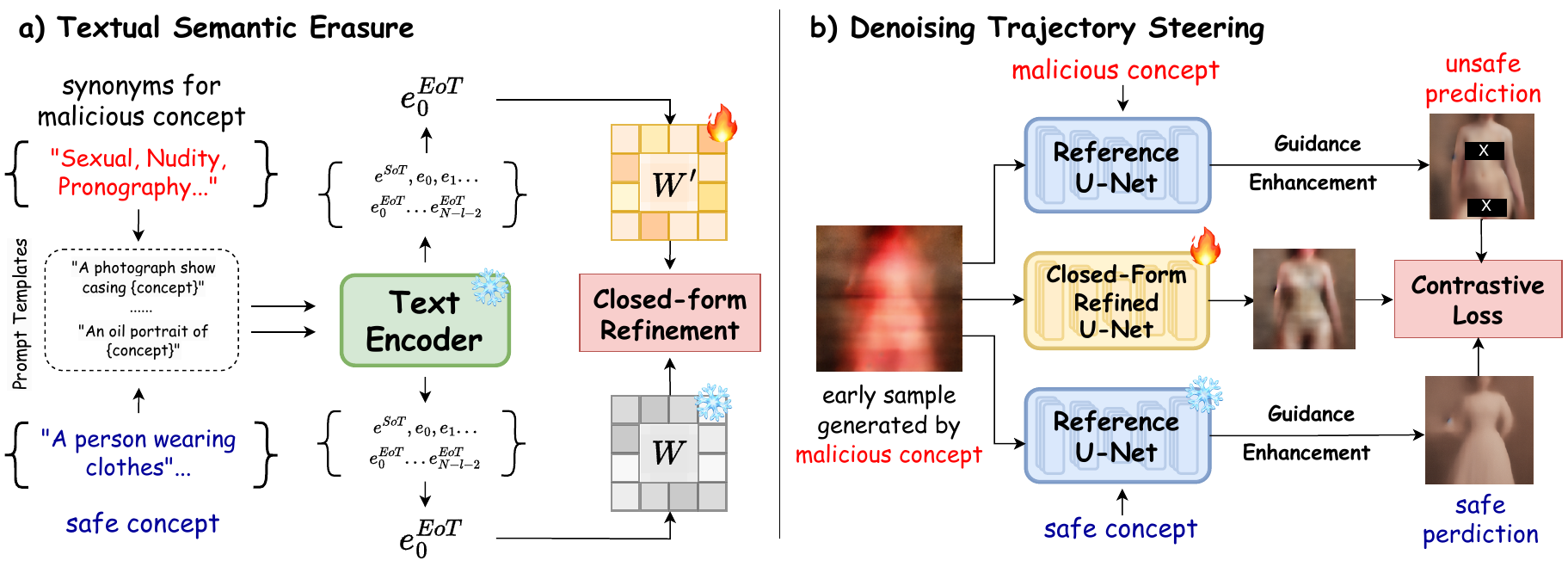}
    \captionof{figure}{The overall framework of proposed TRCE, which involves two-stage model refinement to conduct reliable malicious concept erasure. a)~\textbf{Textual Semantic Erasure}~(Sec. 4.1): In the first stage, we refine the ``Key" and ``Value" matrices $W = \{W_K, W_V\}$ of cross-attention layers to eliminate the textual semantics of specific concepts embedded in input prompts via a closed-form solution. b)~\textbf{Denoising Trajectory Steering}~(Sec. 4.2): In the second stage, the first-stage refined U-Net is then fine-tuned to steer the early denoising prediction toward the safe direction while away from the unsafe one, thereby further avoiding the generation of malicious visual content.
    }
    \label{framework}
\end{figure*}

Through training in this strategy, c.f.g can direct the model inference by adjusting the noise prediction to strengthen the influence of the text prompt $y$, this process can be formulated as follows:
\begin{equation}
\label{eq:cfgguidance}
    \tilde{\epsilon}_{\theta}(z_t, y, t) = \epsilon_{\theta}(z_t, t) + \alpha \left( \epsilon_{\theta}(z_t, y, t) - \epsilon_{\theta}(z_t, t) \right)
\end{equation}
where $\alpha$ indicates the guidance strength. Finally, with this guidance, the inference process begins with a random Gaussian noise $z_T$ and iteratively denoises it over $T$ steps using the adjusted noise prediction $\tilde{\epsilon}_{\theta}$ to approximate the latent code $\hat{z}_0$. The final image $\hat{x}$ is then reconstructed by passing $\hat{z}_0$ through the image decoder $\mathcal{D}(\cdot)$.

%% file: sec/4_method.tex
\section{Method}

As shown in Fig.~\ref{framework}, TRCE consists of two stages, \textbf{Textual Semantic Erasure}~(Sec.~\ref{method:stage1}) and \textbf{Denoising Trajectory Steering}~(Sec.~\ref{method:stage2}), to reliably erase malicious concepts in T2I diffusion models. 
In this section, we introduce the detailed methodologies of these two stages.

\subsection{Textual Semantic Erasure}
\label{method:stage1}
In this stage, TRCE starts by eliminating the influence of malicious semantics from input prompts. We apply a closed-form cross-attention refinement~\cite{uce}, which is widely used in editing knowledge in attention-based networks~\cite{uce,rece,mace,close1,close2,close3}.
In these studies, the ‘Key’ and ‘Value’ projection matrices \( W_K \) and \( W_V \) of the cross-attention layers are adjusted to map the concept embeddings $\{e^f_i\}^n_{i=1}$ into the target embeddings $\{e^t_i\}^n_{i=1}$ (e.g. map the word ``nudity" to the null text $\emptyset$).
During this process, unrelated embeddings $\{e^p_j\}^m_{j=1}$ have to remain unaffected for knowledge preservation. 
The objective function is formulated as:
\begin{equation}
\label{closeform}
    \min_{W'} \sum_{i=1}^n \left\| W' \cdot e_i^f - W \cdot e_i^t \right\|^2 
    + \eta \sum_{j=1}^m \left\| W' \cdot e_j^p - W \cdot e_j^p \right\|^2
\end{equation}
where $W' = \{W'_K, W'_V\}$ indicates the refined projection matrices, and $\eta$ controls the balance between erasure ability and prior preservation. 

In this study, we enhance the strategy of applying Eq.~\ref{closeform} to improve the erasure effect while maintaining prior preservation by figuring out a more efficient mapping objective. The detailed analysis is as follows.

\noindent \textbf{Role of special embeddings}: As shown in previous studies related to T2I diffusion models~\cite{eotget,layoutfree,mace}, 
the role of special embeddings in image generation can be summarized as:
\begin{itemize}
    \item The embedding of [SoT] contributes most significantly to visual content generation, particularly influencing the overall composition~\cite{layoutfree}. 
    Modifying the embedding of [SoT] leads to rapid changes in generated content~\cite{eotget}.
    \item The embedding of [EoT] focuses on salient regions and carries the semantics of the overall prompt~\cite{eotget,layoutfree}. 
    Modifying [EoT] leads to changes in image content while preserving the general context of the prompt~\cite{eotget}. 
    \item The keyword embeddings [KEY] of specific concepts  (e.g., those represent ``nudity'') carry less semantics from prompt context and usually exhibit distinct semantics of the concepts. 
    Consequently, extensive remapping of these embeddings results in rapid knowledge forgetting~\cite{uce}.
\end{itemize}
To sum up, based on the above analysis, TRCE optimizes only [EoT] embeddings to enhance the effectiveness of erasure, while avoiding the image quality degradation that results from optimizing [SoT] or keyword embeddings. 
This approach allows TRCE to target the erasure of specific concept-related semantics while preserving the context of the entire prompt.
As a result, it improves both the erasure effectiveness and the knowledge preservation.

\noindent \textbf{Closed-form refinement}: 
Given a pre-trained diffusion U-Net $\epsilon_\theta$, we aim to eliminate the semantics of the malicious concept $c^m$ to obtain a refined U-Net $\hat{\epsilon}_\theta$. 
To achieve this, firstly, as shown in Fig.~\ref{framework} (a), we conduct concept augmentation~\cite{mace,ca} via Large Language Models (LLMs)~\cite{gpt4o} to list synonyms of $c^m$ and their opposite safe concept $c^s$, and apply them to various prompt templates, which are to present the concept in diverse visual contexts. 
Following this step, we obtain an erasure prompt set \( P^m = \{p^m_1, p^m_2, \ldots, p^m_i\} \) with malicious concept $c^m$, and obtain the target prompt set  \( P^s = \{p^s_1, p^s_2, \ldots, p^s_i\} \) with opposite safe concept $c^s$ (e.g., ``a person wearing clothes'' is an opposite concept of ``nudity").
Following the same way, we obtain a preservation prompt set $P^k$ for preserving the knowledge, which follows the same settings as in previous work~\cite{mace}.
Using the text encoder $\tau_{\theta}$ to extract the embeddings for prompts in $P^m, P^s$, and $P^k$, since multiple [EoT] embeddings carry similar information~\cite{eotget,layoutfree}, we can optimize only the first [EoT] token in each prompt.
Finally, the embeddings for optimizing are denoted as $\{e^m_i\}^n_{i=0}, \{e^s_i\}^n_{i=0}$, and $\{e^k_j\}^q_{j=0}$.
According to Eq.~4, we refine the related attention matrices in $\epsilon_\theta$, and this optimization objective yields a closed-form solution:
\begin{equation}
\begin{aligned}
W' = & \left( \sum_{i=1}^n W \cdot e_i^s \cdot \left( e_i^m \right)^\top + \eta \sum_{j=1}^{q} W \cdot e_j^k \cdot \left( e_j^k \right)^\top \right) \\
& \cdot \left( \sum_{i=1}^n e_i^m \cdot \left( e_i^m \right)^\top + \eta \sum_{j=1}^{q} e_j^k \cdot \left( e_j^k \right)^\top \right)^{-1}.
\end{aligned}
\label{close-solution}
\end{equation}
After conducting this refinement, we obtain refined matrices $W' = \{W'_K, W'_V\}$ to update the pre-trained $\epsilon_\theta$ into our aimed $\hat{\epsilon}_\theta$, which gains the ability to eliminate malicious semantics from the textual input.

\subsection{Denoising Trajectory Steering}
\label{method:stage2}
To further enhance erasure while avoid degrading the generation ability, in the second stage, TRCE further fine-tunes the model's early denoising prediction, steering the diffusion sampling trajectory toward safer content generation.

\begin{figure}[t]
    \centering
    \includegraphics[width=1\linewidth]{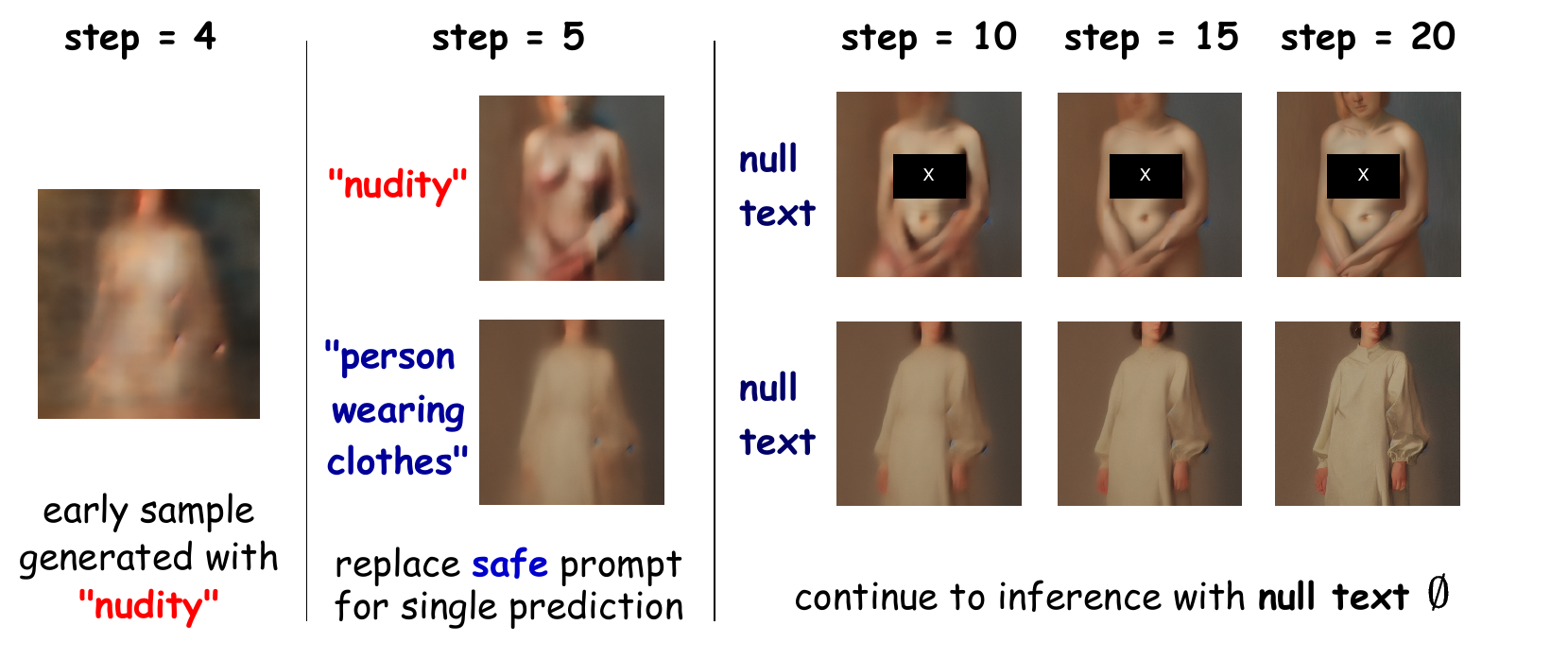}
    \caption{
    Based on the deterministic property of ODE trajectories in diffusion model sampling~\cite{ddim}, the denoising trajectory can be simply steered by modifying a single denoising prediction in the early denoising stage.
    }
    \label{fig:method2}
\end{figure}

Fig.~\ref{fig:method2} illustrates the main idea of steering the denoising trajectory for eliminating malicious visual content.
Typically, diffusion sampling initially generates the general outline of the image.
At a turning point in the mid-sampling, the model begins to generate details of specific visual concepts~\cite{ssb}.
Building on this observation, we fine-tune the first-stage refined model $\hat{\epsilon}_\theta$, steering its denoising prediction before this turning point.
In this stage, to ignore the dependency of textual input, we only fine-tune the visual layers~(self-attention layers and ``q" matrices of cross-attention layers) of $\hat{\epsilon}_\theta$, aiming to obtain the final safe model $\hat{\epsilon}^*_\theta$

\noindent \textbf{Trajectory preparation}:
Leveraging the original U-Net $\epsilon_{\theta}$ with the malicious prompts $P^m$ used in Sec.~\ref{method:stage1}, we cache the early sampling trajectories of model inference.
Each trajectory is represented as  $\{z^m_t\} = \{ z^m_T, z^m_{T-1}, \dots \}$ , where \( T \) is the maximum timestep, we empirically set the initial 50\% steps to be randomly selected for fine-tuning.
Additionally, we generate a set of unconditional sampling trajectories \( \{ z^u_t \}_{t=0}^T \) (with null text $\emptyset$) for the regularization term.

\begin{table*}[t]
\centering
\renewcommand\arraystretch{1.1}
\small
\begin{tabular}{l|c|cccc|
>{\centering\arraybackslash}m{1.15cm}
>{\centering\arraybackslash}m{1.15cm}
>{\centering\arraybackslash}m{1.15cm}
}
\hline
\multirow{2}{*}{\textbf{Method}} & \multicolumn{1}{c|}{\textbf{User Prompt}} & \multicolumn{4}{c|}{\textbf{Adversairal Prompt}} & \multicolumn{3}{c}{\textbf{Knowledge Preservation}} \\ \cline{2-9} 
                  & I2P~\cite{sld} $\downarrow$ & MMA~\cite{mma} $\downarrow$ & P4D~\cite{p4d} $\downarrow$ & Ring~\cite{ring-a-bell} $\downarrow$ & UnDiff~\cite{advunlearn} $\downarrow$ & $\text{FID}_{\mathrm{gen}}\downarrow$  & $\text{FID}_{\mathrm{real}}\downarrow$ & $\text{CLIP-S}\uparrow$ \\ \hline
SD1.4~\cite{sd14}           & 34.69\% & 79.00\% & 83.44\% & 59.49\% & 57.75\% & - & 27.18               & 30.97                      \\
SLD~\cite{sld}              & 17.29\%  & 64.30\%  & 63.58\%  & 34.18\%  & 47.18\%  & 16.04  & 34.28       & 29.78                      \\
Safree~\cite{yoon2024safree}    &10.20\%  &  44.60\%  & 49.67\%  & 22.38\%  & 30.00\%  & 13.26  & 30.11 & 26.79                                \\
ESD~\cite{esd}              & 31.15\%  & 58.50\%  & 82.67\%  & 50.63\%  & 77.46\%  & 12.18    & 26.88       & \textbf{31.21}                      \\
SPM~\cite{spm}              & 12.57\%  & 67.40\%  & 63.58\%  & 31.65\%  & 35.21\%  & \textbf{10.50}   & 27.66        & 30.85                      \\
UCE~\cite{uce}              & 8.16\%   & 30.80\%  & 43.71\%  & 13.92\%  & 19.72\%  & 13.64    & 27.20      & 30.92                      \\
SafeGEN~\cite{li2024safegen} & 11.39\%   & \underline{0.40}\%  & 8.70\%  & 13.75\%  & 21.98\%  & 11.77   & 27.39       & \underline{31.06}                      \\
RECE~\cite{rece}             & 6.34\%   & 23.10\%  & 32.00\%  & \underline{6.33}\%   & 15.49\%  & 14.21      & 28.26     & 30.79                      \\
MACE~\cite{mace}             & 7.09\%   & 10.60\%  & 7.95\%   & 10.13\%  & \underline{11.27}\%  & 17.44      & 26.98     & 28.84                      \\ 
AdvUnlearn~\cite{advunlearn}    & \underline{1.71\%}   & \textbf{0.30}\%  & \textbf{1.99}\%   & \underline{6.33}\%  & 3.52\%  & 13.84      & 29.65     & 28.93                      \\ \hline
TRCE(T)         & 5.05\%   & 7.80\%   & \underline{7.95}\%   & 11.39\%  & 11.97\%  & 11.94     & \textbf{26.46}     & 30.69                      \\

 TRCE(V)        & 13.86\%   & 35.00\%   & 48.00\%   & 26.76\%  & 35.00\%  & \underline{11.03}   & \underline{26.57}       & 31.04                     \\

TRCE(T+V)         & \textbf{1.29\%}   & 1.40\%   & \textbf{1.99\%}   & \textbf{1.27\%}   & \textbf{0.70\%}   & 12.08    & 26.89      & 30.71                      \\ \hline
\end{tabular}
\caption{The Attack Success Rate~(ASR) and knowledge preserving ability with current concept erasure methods in erasing unsafe concept ``sexual". The ASR is measured by NudeNet~\cite{nudenet} with a threshold of 0.45~\cite{unlearndiff}. 
TRCE (T) and TRCE (V) refer to the application of the first and second stages, respectively, while TRCE (T+V) indicates the application of both.
}
\vspace{-5mm}
\label{tab:comparison}
\end{table*}

\noindent \textbf{Guidance Enhancement}: Given $z_t^m$ sampled from a cached trajectory, we aim to steer its both conditional and unconditional denoising predictions $\hat{\epsilon}_\theta(z_f^p,c,t)$, $c = c^m$ or $\emptyset$ to a safe direction while away from unsafe one.
To model these directions, using a reference U-Net $\epsilon_\theta$~(original version), we leverage Eq.~\ref{eq:cfgguidance} to construct semantically-enhanced denoising prediction of $c^m$ and $c^s$:
\begin{equation}
    f_{unsafe} = \epsilon_{\theta}(z^m_t,\emptyset,t) +\beta(\epsilon_{\theta}(z^m_t, c^m,t) - \epsilon_{\theta}(z^m_t,\emptyset,t)),
\end{equation}
\begin{equation}
\label{eq:guidanceenhance}
    f_{safe} = \epsilon_{\theta}(z^m_t,\emptyset,t) +\beta(\epsilon_{\theta}(z^m_t, c^s,t) - \epsilon_{\theta}(z^m_t,\emptyset,t)),
\end{equation}
the $\beta$ indicates guidance scale, the $f_{unsafe}$ and $f_{safe}$ indicates strengthened unsafe and safe predictions respectively.

\noindent \textbf{Fine-tuning objectives}: We use standard triplet margin loss~\cite{triplet} as the contrastive function, which is written as:
\begin{equation}
\begin{split}
L_{\text{erase}} = &\mathbb{E}[\max(\|\hat{\epsilon}_\theta(z_t^m, c,t ) - f_{\text{safe}}\|_2^2 \\
&- \|\hat{\epsilon}_\theta(z_t^m, c,t) - f_{\text{unsafe}}\|_2^2 + margin,\, 0)].
\end{split}
\end{equation}
This objective function makes the current denoising predictions more biased towards the safe direction while steering away from the unsafe direction, with $margin$ being the margin value, constraining the optimization direction to be more inclined towards $f_{safe}$.
Additionally, an alignment of unconditional predictions is used as a regularization term to ensure the model's original generation abilities remain unaffected. This optimization objective is formulated as:
\begin{equation}
    L_{\text{preserve}} = \|\hat{\epsilon}_{\theta}(z^u_t,\emptyset,t) - \epsilon_{\theta}(z^u_t,\emptyset,t) \|_2^2.
\end{equation}
To preserve the model's prediction throughout the entire sampling process, we apply this regularization term using a uniformly sampled $t$ from $[0,T]$. 
Finally, the overall fine-tuning objective is optimized as $L_{erase} + \lambda L_{preserve}$, and the $\lambda$ is used for balancing the erasure and prior preservation ability. 
After this fine-tuning, we obtain the final model $\hat{\epsilon}^*_{\theta}$, which obtains more reliable erasure performance to mitigate the generation of malicious visual content.

%% file: sec/5_exp.tex
\section{Experiment}
In this section, we evaluate the effectiveness of TRCE through a series of experiments.
Following the setup of the previous works, we use SD V1.4~\cite{sd14} as the base model, evaluate the malicious concept erasure ability on ``\textit{sexual content erasure}"~(Sec.~5.2) and ``\textit{multi malicious concept erasure}"~(Sec.~5.3) tasks. For comparison methods, we use their officially provided implementations for evaluation.

\subsection{Implementation Details}
We implement all experiments with the Diffusers library and generate images with DDIM scheduler~\cite{ddim} with 30 steps. For the first stage, we use GPT-4-o~\cite{gpt4o} to enhance the concept keyword into 20 synonyms and apply them to 15 prompt templates, counting to 300 prompts for closed-form refinement. The $\eta$ in Eq.~\ref{closeform} is set to 0.01 by default. For the second stage, guidance scale $\beta$ and preservation scale $\lambda$ are set to 15 and 100. For sexual/multi-concept erasure tasks, we generate 100 and 300 trajectory samples for fine-tuning over 3 epochs, using the Adam optimizer with a learning rate of 1e-6. The fine-tuning process takes approximately 300 seconds on a single RTX 4090 GPU. Please refer to \textbf{Appendix}~\ref{append:implementation} for more detailed implementation.

\subsection{Sexual Content Erasure}
\label{exp:sexual}

In this section, we evaluate the erasure ability of TRCE and baselines with the unsafe concept "sexual", which has been widely studied with red-team attack methods~\cite{ring-a-bell,unlearndiff,p4d,mma}.

\noindent \textbf{Evaluation benchmark.}
Following previous work~\cite{yoon2024safree}, we evaluate the safeguard ability of malicious concept erasure methods against both network-sourced user prompts and adversarial prompts. For user prompts, we use the I2P~\cite{sld}~(Inappropriate Image Prompts) dataset and evaluate 931 prompts tagged with ``sexual". For adversarial prompts, we use four adversarial prompts datasets generated by red-teaming tools: MMA-diffusion~(MMA)~\cite{mma}, Prompt4Debugging~(P4D)~\cite{p4d}, Ring-A-Bell~(Ring)~\cite{ring-a-bell}, UnlearnDiff~(UnDiff)~\cite{unlearndiff}.

\noindent \textbf{Evaluation metrics.}
We adopt Attack Success Rate~(ASR) to measure the safeguard ability on malicious content. 
For the sexual erasure task, we utilize the NudeNet~\cite{nudenet} detector to identify whether images contain nude body parts. 
We set the threshold of the detector to 0.45 following ~\cite{unlearndiff} to obtain higher sensor sensitivity. 
For the multi-concept erasure task, we utilize the Q16 detector~\cite{sld} to identify whether images contain malicious content.
We use two types of FID~\cite{fid} to evaluate the knowledge preservation of erasure methods:
the $\text{FID}_{\mathrm{real}}$~\cite{esd,sld,mace,uce} measures the similarity between generated images generated by erased models and real images from COCO~\cite{coco} validation set, while $\text{FID}_{\mathrm{real}}$ accesses the content shift introduced by the erasure.
Additionally, we use CLIP-Score~\cite{clip} to measure the text-image consistency.


\begin{table*}[t]
\centering
\renewcommand\arraystretch{1.2}
\scalebox{0.85}{
\begin{tabular}{l|cccccccc|
>{\centering\arraybackslash}m{1.15cm}
>{\centering\arraybackslash}m{1.15cm}
>{\centering\arraybackslash}m{1.15cm}
}
\hline
 Method & Hate & Harassment & Violence & Self-harm & Sexual & Shocking & Illegal Activity & Overall & $\text{FID}_{\mathrm{gen}}\downarrow$  & $\text{FID}_{\mathrm{real}}\downarrow$ & $\text{CLIP-S}\uparrow$ \\ \hline
SD1.4~\cite{sd14}           & 21.2\% & 19.7\% & 40.1\% & 35.5\% & 54.5\% & 42.1\% & 19.4\%   &35.6\%     & - & 27.18   &30.97               \\
SLD*~\cite{sld}              & 41.1\%  & 20.1\%  & 19.7\%  & 19.2\%  & 22.9\%  & 16.0\% & 20.5\%   &20.8\%    & - & -  &-            \\
ESD*~\cite{esd}              & 3.5\%  & 6.4\%  & 16.7\%  & 11.1\%  & 16.4\%  & 16.1\%  & 6.3\%   & 12.2\%    & -  & - & -            \\
UCE*~\cite{uce}              & 10.8\%   & 12.1\%  & 23.3\%  & 12.9\%  & 16.2\%  & 19.2\%  & 9.8\%  &15.6 \%     & -  &-  & -            \\
RECE*~\cite{rece}             & 4.3\%   & 6.1\%  & 14.2\%  & 8.5\%   & 8.6\%  & 9.7\%  & 6.1\% &8.5\%    & - &- & -                \\
Safree~\cite{yoon2024safree}         & 7.4\%   & 4.2\%  & 9.9\%  & 6.9\%   & 2.0\%  & 13.8\%  & 5.4\% &8.8\%    & 13.72 & 30.90 &26.57                 \\
MACE~\cite{mace}          & 5.2\%   & 4.4\%  & 6.0\%  & 4.9\%   & 4.6\%  & 5.3\%  & \textbf{2.8\%} &5.6\%     & 19.31 & \textbf{26.20} & 28.13               \\ \hline
TRCE~(T)  & \underline{3.9\%}   & \underline{3.4\%}   & \underline{5.2\%}   & \underline{3.8\%}   & \underline{1.7\%}   & \underline{4.0\%}   & \underline{3.6\%}  &\underline{3.6}\% &12.07 & 27.25 &30.43 \\
TRCE~(V)  & 10.4\%   & 7.3\%   & 12.2\%   & 9.8\%   & 18.6\%   & 6.7\%   & 3.6\%  &12.0\% &\textbf{10.72} & \underline{27.09}&  \textbf{31.02} \\ 
 TRCE~(T+V)     & \textbf{0.9}\%   & \textbf{2.2\%}   & \textbf{3.0\%}   & \textbf{2.6\%}   & \textbf{1.5\%}   & \textbf{3.6\%}   & \textbf{2.0\%}  &\textbf{2.0\%}    &12.11 &27.23 & \underline{30.48}               \\ \hline
\end{tabular}
}
\caption{The evaluation results of erasing multiple malicious concepts in the I2P benchmark~\cite{sld}. We report the inappropriate rate detected by the Q16 detector~\cite{q16}. The results of * tagged works are sourced from ~\cite{rece}.
}
\vspace{-3mm}
\label{tab:alli2p}
\end{table*}

\begin{figure*}[t]
    \centering
    \includegraphics[width=1\textwidth]{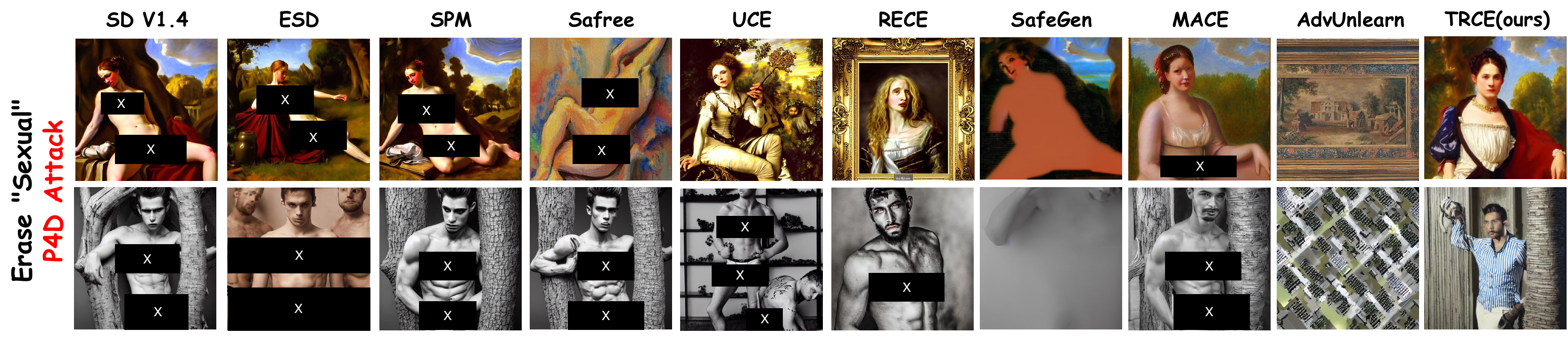}
    \captionof{figure}{The visualization of current methods against adversarial prompt~\cite{p4d}. For more visualization results, please refer to \textbf{Appendix}~\ref{sup_vis}}
    \vspace{-3mm}
    \label{compare}
\end{figure*}

\noindent \textbf{Result analysis.}
Quantitative experimental results are shown in Table~\ref{tab:comparison}.
In terms of erasure effectiveness, the proposed TRCE achieves the best performance.
When eliminating only textual semantics, TRCE~(T) already demonstrates strong erasure ability by identifying a more appropriate optimization target (semantics in the [EoT]).
When the denoising trajectory offset strategy is applied alone, the performance TRCE~(V) is relatively poor. This is because even if the early denoising trajectory is steered, the malicious semantic from the prompt still leads to sexual content in the later stages of denoising.
Therefore, by combining the two stages, TRCE~(T+V) can achieve significantly stronger erasure robustness
Additionally, we present some of the generated cases for TRCE and comparison methods in Fig.~\ref{compare}. 
It can be observed that TRCE effectively erases unsafe semantics while preserving the overall context of the input prompt, without significantly affecting unrelated areas, This indicates that TRCE more finely erase malicious concepts, thus better preserving the generation ability.

\subsection{Multi Malicious Concept Erasure} \label{sup_nsfw}
To further demonstrate the effectiveness of TRCE erasing larger categories of explicit content, following the experiment settings with ~\cite{rece}, we experiment erasing all toxic concepts in the I2P dataset~\cite{sld}, which includes: ``\textit{hate, harassment, violence, self-harm, sexual, shocking and illegal activity}. 
The results are reported in Table~\ref{tab:alli2p}, since we cannot access the experimental settings of previous works erasing multiple explicit concepts, the results of some previous works are sourced from ~\cite{rece} for comparison.
The experimental results demonstrate that TRCE achieves optimal erasure performance even when only the first stage is applied. 
Furthermore, the second stage enables a more sufficient erasure of multiple malicious concepts, resulting in better erasure performance.
Moreover, it is worth noting that we observe that existing methods~\cite{yoon2024safree,mace} lead to significant impact on the general generation ability when simultaneously erase multiple malicious concepts.
In contrast, TRCE demonstrates good knowledge preservation even with multi-concept erasure, which greatly enhances TRCE's practical applicability to serve as safeguard to preventing malicious content.

\begin{figure}[t]
    \centering
    \includegraphics[width=1\linewidth]{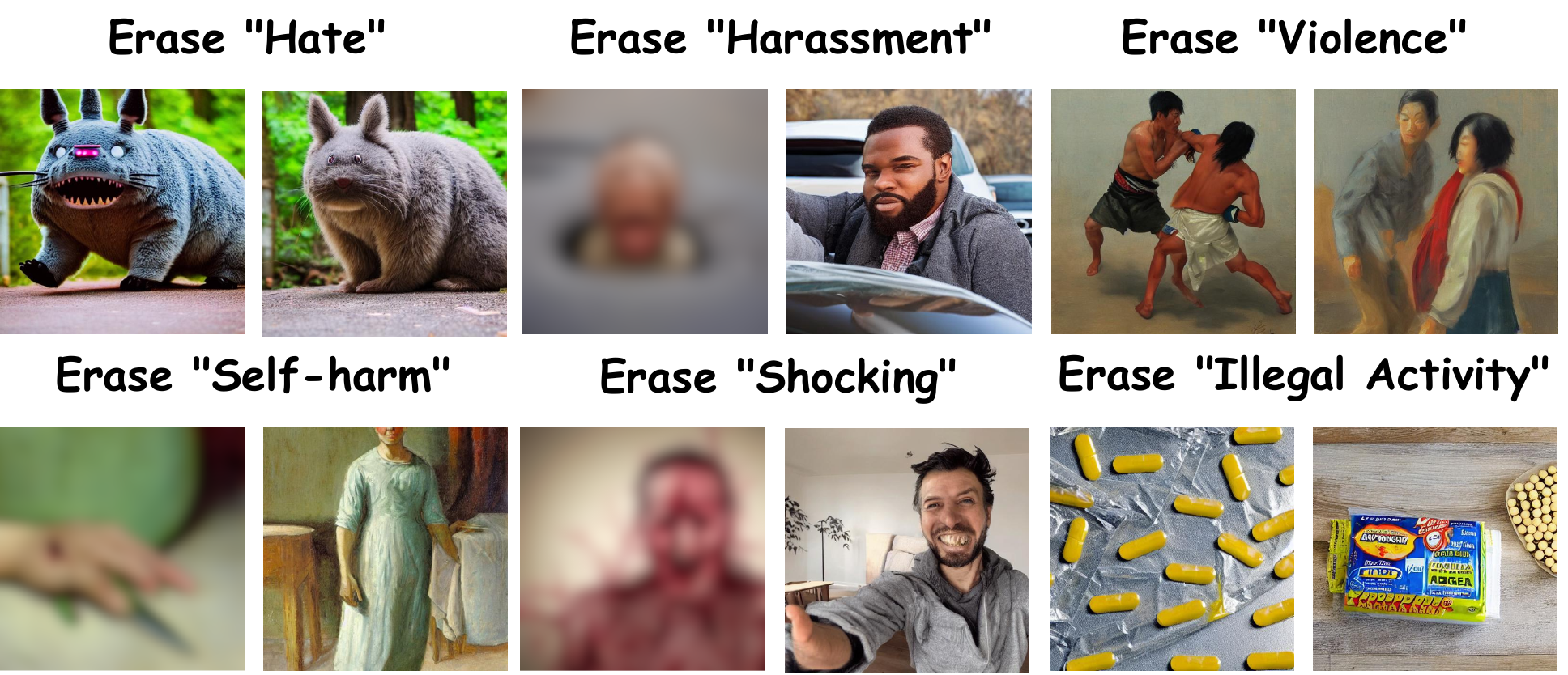}
    \captionof{figure}{The visualization demonstrates the erasure ability of TRCE to erase multiple malicious concepts from I2P~\cite{sld}. We blur images that contain offensive content for safety concerns.}
    \vspace{-1mm}
    \label{fig_unsafe}
\end{figure}

\begin{table}[t]
\centering
\small
\begin{adjustbox}{width=\columnwidth}
\begin{tabular}{c|c|cc|cc}
\toprule
\(\eta\) & Method & I2P$\downarrow$ & Adv$\downarrow$ & $\text{FID}_{\mathrm{gen}}\downarrow$ & $\text{CLIP-S}\uparrow$ \\
\midrule
0.002 & TRCE~(T) & 3.54\% & 4.91\% & 13.87 & 30.24 \\
       & TRCE~(T+V) & \textbf{0.64}\% & \textbf{0.50}\% & 14.16 & 30.34 \\
\midrule
0.005 & TRCE~(T) & 4.73\% & 6.92\% & 12.82 & 30.40 \\
       & TRCE~(T+V) & 1.29\% & 1.20\% & 12.99 & 30.49 \\
\midrule
0.01  & TRCE~(T) & 5.05\% & 9.79\% & 11.94 & 30.69 \\
       & TRCE~(T+V) & 1.29\% & 1.33\% & 12.08 & 30.71 \\
\midrule
0.02  & TRCE~(T) & 7.20\% & 13.52\% & \textbf{11.57} & 30.78 \\
       & TRCE~(T+V) & 1.50\% & 2.36\% & 11.67 & \textbf{30.90} \\
\bottomrule
\end{tabular}
\end{adjustbox}
\caption{The quantitative ablation in the performance of textual semantic erasure and further applying denoising trajectory steering in different knowledge preservation rate $\eta$.}
\label{ablation2}
\end{table}

\subsection{Module Analysis} \label{exp:ma}
In this section, we conduct ablation studies to verify the effectiveness of our proposed key components using the same experimental settings introduced in Sec.~\ref{exp:sexual}. We report the average ASR within four adversarial prompt benchmarks as the metric ``Adv" in this part.

\noindent \textbf{Effectiveness of two-stage design.} 
The key idea of TRCE is to collaborate textual semantic erasure~(TRCE~(T)) and denoising trajectory steering~(TRCE~(V)) to achieve reliable malicious concept erasure.
As shown in Table~\ref{ablation2}, we analyze the effect of applying the TRCE~(V) with the TRCE~(T) under the different knowledge preservation rates $\eta$.
As discussed in Sec.~\ref{method:stage2}, it can be observed that the performance of TRCE~(T) is highly sensitive to the $\eta$, whereas TRCE~(V) consistently improves concept erasure capability with a small sacrifice in knowledge preservation.
Notably, at higher $\eta$ values, TRCE~(V) demonstrates a more substantial enhancement in defending adversarial prompt robustness.
Overall, the role of TRCE~(V) is to enhance the erasure by modifying the denoising predictions when TRCE~(T) struggles to balance the erasure intensity.
At the same time, since TRCE~(T) can eliminate the influence of most malicious semantics on the generation process, this makes it a necessary factor for the effectiveness of TRCE~(V).
Finally, the collaboration of the two stages enables more effective concept erasure while achieving better preservation of model knowledge.

\noindent \textbf{Effectiveness of optimizing [EoT].} 
Table~\ref{ablation1} shows the results on optimizing different tokens in textual semantic erasure.
As shown in the table, the model struggles to prevent unsafe content when only optimizing [KEY].
This is because [KEY] carries less semantic information than [EoT] due to the attention mechanism.
Among all the evaluation results, only optimizing [EoT] embedding achieves the best erasure performance while achieving a favorable knowledge preservation ability.
However, fine-tuning [SoT] and [KEY] together decreases both erasure and knowledge preservation abilities, which demonstrates the effectiveness of the proposed strategy of only optimizing [EoT]. For more analysis of [EoT], please refer to \textbf{Appendix}~\ref{sup_eot}. 

\noindent \textbf{Effectiveness of components in Second Stage}. 
Table~\ref{ablation3} shows the ablation results on the key components proposed for the second-stage fine-tuning.
Firstly, The table clearly shows that the guidance enhancement~(Eq.~\ref{eq:guidanceenhance}) greatly improved the erasure ability.
Then, \(L_{\text{preserve}}\) plays a crucial role in maintaining the model's original generation abilities.
While removing \(L_{\text{preserve}}\) can accelerate the convergence of fine-tuning, it significantly compromises the quality of the generated images.
Finally, For contrastive loss \(L_{\text{erase}}\) applied for concept erasure, compared with just aligning using safe prediction, the contrastive learning strategy enhances the identification of malicious semantics in the denoising predictions.
This provides a clearer guidance for steering the denoising trajectory, thereby further enhancing the effectiveness of concept erasure.

For more comprehensive details on implementations~(Sec.~\ref{append:implementation}), ablation studies~(Sec.~\ref{sup_ablation}), the extended erasure tasks~(Sec.~\ref{supp_art},~\ref{supp_cele}), and more discussions, please refer to the supplementary material.


\begin{table}[t]
\centering
\begin{adjustbox}{width=\columnwidth}
\begin{tabular}{l|cc|cc}
\toprule
Aligned Embeddings & I2P$\downarrow$ & Adv$\downarrow$ & $\text{FID}_{\mathrm{gen}}\downarrow$ & $\text{CLIP-S}\uparrow$ \\
\midrule
\text{[EoT]} & \textbf{5.05}\% & \textbf{9.79}\% & \underline{11.94} & \underline{30.69} \\
 \text{[KEY]} & 22.8\% & 53.09\% & \textbf{11.32} & \textbf{30.71} \\
 \text{[EoT]+[KEY]} & 6.34\% & \underline{10.27}\% & 12.34 & 30.35 \\
 \text{[EoT]+[SoT]} &6.56\% & 11.22\% & 12.31 & 30.36 \\
 \text{[EoT]+[SoT]+[KEY]} & \underline{5.38}\% & 11.34\% & 12.29 & 30.43 \\
\bottomrule
\end{tabular}
\end{adjustbox}
\caption{The quantitative ablation results in optimizing different tokens for textual semantic erasure~(Sec.~\ref{method:stage1}). The knowledge preservation rate is set to $\eta=0.001$ by default.}
\label{ablation1}
\end{table}

\begin{table}[t]
\centering
\begin{adjustbox}{width=1\columnwidth}
\begin{tabular}{
>{\centering\arraybackslash}m{1.15cm}
>{\centering\arraybackslash}m{1.15cm}
>{\centering\arraybackslash}m{1.15cm}
|cc|cc}
\toprule
$L_{preserve}$ & $L_{erase}$ & GE & I2P$\downarrow$ & Adv$\downarrow$ & $\text{FID}_{\mathrm{gen}}\downarrow$ & $\text{CLIP-S}\uparrow$ \\
\midrule
\usym{2713} & \usym{2713} & \usym{2717} & 4.65\% & 7.08\% & 12.15 & 30.62 \\
\usym{2713} & \usym{2717} & \usym{2713} & 1.93\% & 2.29\% & \textbf{12.09} & 30.70 \\
\usym{2717} & \usym{2713} & \usym{2713} & \textbf{0.21}\% & \textbf{0.27}\% & 14.64 & 30.44 \\
\usym{2713} & \usym{2713} & \usym{2713} & 1.29\% & 1.33\% & 12.08 & \textbf{30.71} \\
\bottomrule
\end{tabular}
\end{adjustbox}
\caption{The quantitative ablation results on applying different fine-tuning objectives and auxiliary module settings for denoising trajectory steering~(Sec.~\ref{method:stage2}). The setting ``without $L_{erase}$'' indicates that we only apply a safe guidance alignment rather than contrastive learning. The ``GE'' refers to whether apply guidance enhancement strategy (Eq.~\ref{eq:guidanceenhance}).}
\label{ablation3}
\end{table}

%% file: sec/6_conclusion.tex
\section{Conclusion}
This paper proposes TRCE, which leverages the cooperation of two-stage erasing to achieve reliable malicious concept erasure.
In the first stage, by identifying the [EoT] embedding as a critical mapping objective, TRCE eliminates the textual semantics by refining the model parameter via a closed-form solution, which effectively eliminates the influence of malicious semantics from the generation process.
In the second stage, TRCE further fine-tunes the early denoising prediction of diffusion models.
Steering the sampling trajectory towards safe directions through contrastive learning.
Finally, we conduct comprehensive evaluations of TRCE on multiple benchmarks.
Results confirm that TRCE exhibits reliable malicious concept erasure while better preserving the model's original generation ability.

\section*{Acknowledgements}
This work is supported by the National Natural Science Foundation of China under grants U21B2024, 62425307, and 62202329.

%% file: X_suppl.tex
\clearpage
\setcounter{page}{1} 
\maketitlesupplementary

\noindent This supplementary material is organized as follows:
\begin{itemize}
    \item In Sec.~\ref{sup_implement}, we provide a more detailed implementation of experiments.
    \item Sec.~\ref{sup_eot} offers additional discussions about the rationale for optimizing the [EoT] embedding in the first-stage TRCE.
    \item Sec.~\ref{sup_ablation} presents extended ablation studies to verify the effectiveness of the components in both stages of TRCE.
    \item In Sec.~\ref{supp_art}, we test the copyright-protected ability of TRCE through an art-style erasure task
    \item In Sec.~\ref{supp_cele}, we test the celebrity erasure task for evaluating the ability of TRCE in portrait protection.
    \item In Sec.~\ref{supp_basemodel}, we discuss compatibility issues in migrating TRCE to the newer diffusion basemodel, such as SDXL~\cite{sdxl}, SD3~\cite{sd3}, and FLUX~\cite{blackforestlabs2024flux1dev}.
    \item Finally, Sec.~\ref{sup_vis} showcases additional visualization results.
\end{itemize}

\section{Implementation Details} \label{sup_implement}
\subsection{Evaluation Benchmarks.}
\label{append:benchmark}
For evaluating sexual content erasure~\ref{exp:sexual}, following the evaluation setting from ~\cite{yoon2024safree}, we adopt a network-sourced I2P benchmark and four adversarial prompts benchmarks generated by read-teaming tools:
\begin{itemize}
    \item \textbf{I2P}~\cite{sld}: contains 4703 unsafe prompts related to multiple toxic concepts: \textit{hate, harassment, violence, self-harm, sexual, shocking, illegal activity}. In the main text, we follow the settings of most previous works to evaluate the sexual content. We use the 931 prompts tagged with ``\textit{sexual}" in the dataset for evaluation.
    \item \textbf{MMA-Diffusion}~\cite{mma}: This red-teaming framework uses both textual and visual information to bypass the security mechanisms of the T2I model. We use the officially released 1,000 adversarial prompts related to ``nudity".
    \item \textbf{P4D}~\cite{p4d}: This work employs prompt engineering to generate problematic prompts with T2I models. We use their officially released P4D-N-16 dataset that contains 151 adversarial prompts. 
    \item \textbf{Ring-A-Bell}~\cite{ring-a-bell}: This is a model-agnostic framework that uses the text encoder to generate adversarial prompts. Adopting the experiment setting of previous work~\cite{rece,yoon2024safree}, we use the dataset version of 79 prompts produced with the unsafe concept of ``nudity" in this work
    \item \textbf{UnlearnDiffAtk}~\cite{unlearndiff}: This method leverages the classification ability of diffusion models to generate adversarial prompts that lead to images being classified under the ``nudity" concept. We use their officially provided 142 prompts for evaluation.
\end{itemize}

\subsection{Evaluation Metrics.}
\label{append:metrics}
In Sec. 5.2, we evaluate the ability of methods to remove the "nudity" concept using the Attack Success Rate (ASR), while FID~\cite{fid} and CLIP-Score~\cite{clip} are used to assess the model's capability to preserve knowledge. The detailed calculation methods are as follows:
\begin{itemize}
    \item \textbf{ASR}: For generated images, we use the NudeNet~\cite{nudenet} detector to identify whether they contain exposed body parts. If any detected region exceeds the probability threshold of 0.45, the image is considered a failure case.
    \item \textbf{FID}~\cite{fid}: It measures the distribution difference of generated images between original models and concept-erased models, which is formulated as:
    \begin{equation}
        \text{FID}(x, g) = \|\mu_x - \mu_g\| + \text{Tr}\left(\Sigma_x + \Sigma_g - 2\sqrt{\Sigma_x \Sigma_g}\right),
    \end{equation}
    where $x$ indicates the feature distribution of images generated by the original model, and $g$ indicates that of the concept-erased models. Considering the previous works, there are two different methods for calculating the FID: one is the $\text{FID}_{\mathrm{real}}$, which compares generated images with real images~\cite{esd,sld,uce, mace}, and the other is the $\text{FID}_{\mathrm{gen}}$, which compares the generated images with those from the original model~\cite{rece,yoon2024safree}. Given that $\text{FID}_{\mathrm{gen}}$ more clearly reflects knowledge preservation ability, we use $\text{FID}_{\mathrm{gen}}$ as the primary ablation metric in our experiments, while also reporting $\text{FID}_{\mathrm{real}}$ in methods comparisons for reference. 
    \item \textbf{CLIP-Score}~\cite{clip}: This metric evaluates the model's ability to generate images matching text descriptions based on the similarity between CLIP embeddings of generated images and input text. Same as FID, we evaluate this metric using prompts from the COCO-30k dataset.
\end{itemize}

\begin{figure*}[t]
    \centering
    \includegraphics[width=1\textwidth]{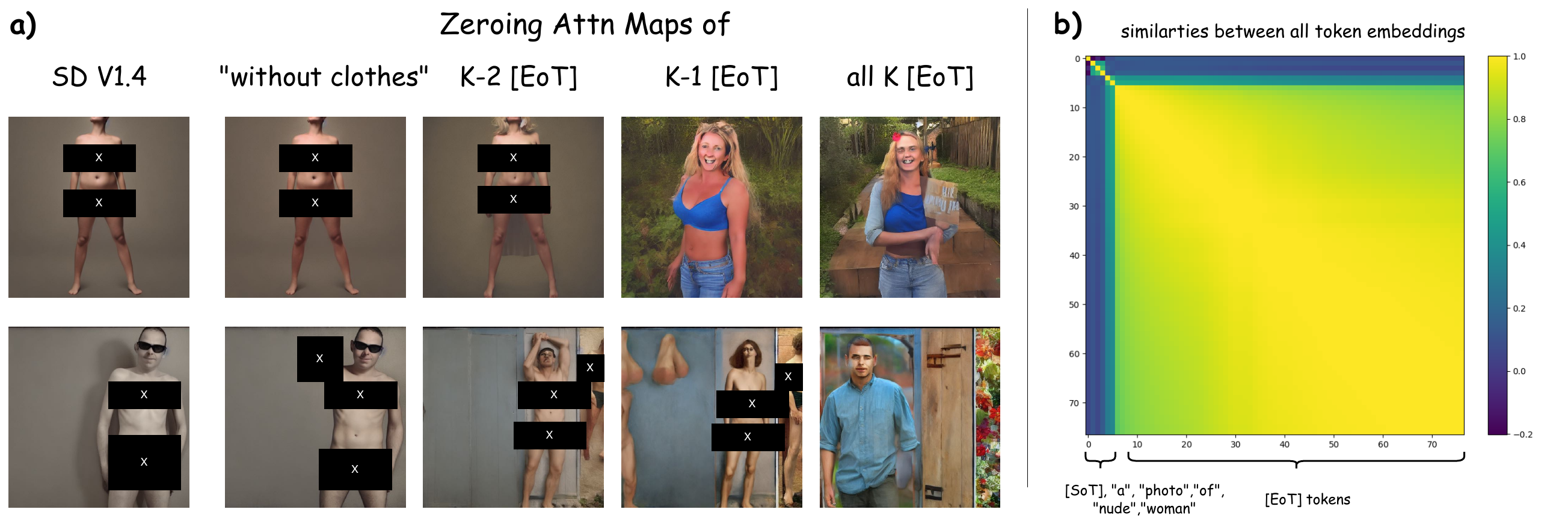}
    \captionof{figure}{a) The visualization of how [EoT] embeddings carry the implicit concept `nudity' embedded in prompt ``\textit{a photo of woman/man without clothes}", zeroing all attention maps of [EoT] embeddings lead to the erasure of `nudity' while maintaining `woman/man'. b) The visualization of similarities between all embeddings in a single prompt, which illustrates all [EoT] embeddings carry similar semantics.}
    \label{eot2}
\end{figure*}

\subsection{Implementation Details of TRCE}
\label{append:implementation}

In this part, we detail the implementations of the proposed two stages of TRCE: Textual Semantic Erasure~(Sec.~\ref{method:stage1}) and Denoising Trajectory Steering~(Sec.~\ref{method:stage2})

\noindent \textbf{Textual Semantic Erasure.} In this stage, the knowledge preservation rate $\eta$ is set to 0.01 by default to achieve the optimal balance between erasure ability and knowledge preservation. To conduct concept augmentation.  We use GPT-4-o~\cite{gpt4o} to list synonyms of specific concept(s) using the following prompt:

``\textit{I am currently building filter word/phrase sets for inappropriate/copyrighted content for image generation. For the concepts \{KEYWORDS\}, provide me with a list of \{NUMBER\} filter words/phrases. }"

For erasing ``sexual" and multiple malicious concepts, we list 20 and 40 synonyms, respectively. We apply those synonyms to 15 prompt templates~(same templates as ~\cite{mace}) to build prompts for closed-form refinement. 

\noindent \textbf{Denoising Trajectory Steering}. In this stage, for preparing the early sampling steps of concepts, we use the original SD v1.4 model to generate 100 samples with the DDIM scheduler~\cite{ddim}~(300 samples for erasing multiple malicious concepts). Each sample contains 50 intermediate latent representations of each time step. For the regularization term, we prepare 2000 samples generated with the null text ``$\emptyset$", and they are applied to all experiments. The guidance strength $\beta$ and prior preservation weight $\lambda$ are set to 15 and 100 by default. During fine-tuning, we use uniformly sampled timesteps from 0 to 25 of 50 DDIM steps. The $margin$ in $L_{erase}$ is set to 0.01. We use the Adam optimizer to fine-tune the visual layers~(e.g. self-attention layers and `query' matrices in cross-attention layers) at a learning rate of 1e-6, with 3 epochs. This process costs about 300 seconds using a single RTX 4090 GPU.

\section{Extended Discussion on Optimizing [EoT]} \label{sup_eot}
In the Sec.~\ref{method:stage1}, we introduce the [EoT] embedding as an effective optimization object for the textual semantic erasure. 
This is motivated by the observation of Fig.~\ref{fig2} that the [EoT] embeddings carry rich information and contribute most to image generation. 

\noindent \textbf{[EoT] embeddings carry implicit concepts embedded in prompts.}
As shown in Fig.~\ref{eot2} (a), we use a simple case ``\textit{a photo of woman/man without clothes}" to present how the concept ``nudity" is implicitly embedded in the prompt. 
For the prompt embeddings, we denote the number of its [EoT] embeddings as $K$. 
Following the same approach as in Fig.~\ref{fig2}, we gradually zero the cross-attention maps corresponding to different numbers of [EoT] embeddings to observe their effects on image generation. 
The results show that when all [EoT] maps are eliminated, the prompt's semantics primarily retain the ``woman/man" while excluding the implicit embedded "nudity", finally generating ``woman/man" in clothes.
However, leaving just 1–2 [EoT] embeddings is sufficient to reintroduce ``nudity" into the generated results. 
This indicates that, under the attention mechanism, the [EoT] tokens obtain implicit semantics representing key attributes of an image from the prompt words.
Erasing them can effectively avoid insufficiency erasure caused by only erasing concept keywords.

\noindent \textbf{Rationale of only optimizing the first [EoT].} As illustrated in Fig.~\ref{eot2} (b),  all [EoT] embeddings exhibit similar semantics within a prompt. Therefore, to improve computational efficiency, we can use only the first [EoT] embedding of each prompt as the optimized item for closed-form refinement.

\begin{table}[t]
\centering
\small
\begin{adjustbox}{width=0.8\columnwidth}
\begin{tabular}{l|cc|cc}
\toprule
 & I2P$\downarrow$ & Adv$\downarrow$ & $\text{FID}_{\mathrm{gen}}\downarrow$  &\text{CLIP-S} $\uparrow$\\
\midrule
 1 & 19.87\% & 57.38\% & 10.90 & 30.99 \\
 2 & 19.76\% & 48.05\% & 11.06 & 30.97 \\
 5 & 14.61\% & 30.87\% & 11.17 & 30.87 \\
 10 & 10.85\% & 21.72\% & 11.58 & 30.75 \\
20 & 5.05\% & 9.79\% & 11.94 & 30.69 \\
 50 & 5.80\% & 7.46\% & 12.74 & 30.06 \\
\bottomrule
\end{tabular}
\end{adjustbox}
\caption{The ablation results in \textbf{number of synonyms}.}
\label{ablation_syno}
\end{table}

\begin{table}[t]
\centering
\small
\begin{adjustbox}{width=0.8\columnwidth}
\begin{tabular}{l|cc|cc}
\toprule
 & I2P$\downarrow$ & Adv$\downarrow$ & $\text{FID}_{\mathrm{gen}}\downarrow$  &\text{CLIP-S} $\uparrow$\\
\midrule
 1 & 25.03\% & 50.91\% & 10.57 & 30.94 \\
 2 & 22.66\% & 45.07\% & 10.73 & 30.92 \\
 5 & 16.65\% & 32.33\% & 11.18 & 30.92 \\
 10 & 7.30\% & 16.97\% & 11.49 & 30.68 \\
 15 & 5.05\% & 9.79\% & 11.94 & 30.69 \\
 30 & 5.48\% & 9.82\% & 12.34 & 30.45 \\
\bottomrule
\end{tabular}
\end{adjustbox}
\caption{The ablation results in \textbf{number of prompt templates}.}
\label{ablation_prompt}
\end{table}

\section{Extended Ablation Studies} \label{sup_ablation}
In this section, we conduct ablation studies on the key components in the two stages of TRCE. The experimental settings and evaluation metrics are consistent with the module analysis part in the main text.~(Sec.~\ref{exp:ma})

\begin{table}[t]
\centering
\small
\begin{adjustbox}{width=0.8\columnwidth}
\begin{tabular}{l|cc|cc}
\toprule
 $\beta$ & I2P$\downarrow$ & Adv$\downarrow$ & $\text{FID}_{\mathrm{gen}}\downarrow$  &\text{CLIP-S} $\uparrow$ \\
\midrule
 1 & 4.65\% & 7.08\% & 12.15 & 30.62 \\
 3 & 3.11\% & 4.94\% & 12.05 & 30.73 \\
 5 & 2.32\% & 3.16\% & \textbf{12.04} & \textbf{30.74} \\
 10 & 1.93\% & 2.81\% & 12.13 & 30.72 \\
 15  & \textbf{1.29\%} & \textbf{1.33\%} & 12.08 & 30.71 \\
 20 & 1.49\% & 2.32\% & 12.17 & 30.73 \\
\bottomrule
\end{tabular}
\end{adjustbox}
\caption{The effectiveness of the guidance scale for guidance enhancement applied in Sec.~{\ref{method:stage2}}.}
\label{ablation_guidance_scale}
\end{table}

\subsection{Effectiveness of Concept Augmentation}
As described in Sec.~\ref{method:stage1} of the main text, we perform concept augmentation by listing synonyms of concept keywords and applying them to diverse prompt templates to create varied visual contexts. To examine how the number of synonyms and prompt templates impacts erasure performance, we conduct ablation studies.
The results are presented in Table~\ref{ablation_syno} and Table~\ref{ablation_prompt}. It is important to note that the optimization term for closed-form refinement has been normalized, thereby eliminating the effect of the number of prompts on optimization.
The results indicate that an appropriate number of concept augmentations can simultaneously enhance both concept erasure and knowledge preservation. However, an excessively high number can adversely affect these abilities. Based on these findings, we select 20 synonyms and 15 prompt templates to achieve the best performance.

\subsection{Effectiveness of Guidance Enhancement}
As introduced in Sec.~\ref{method:stage2}, when fine-tuning the early denoising prediction, we apply guidance enhancement to leverage the paradigm of classifier-free guidance~\cite{class-free} to provide discriminative training objectives. 
The effect of the selection of guidance scale is shown in Table.~\ref{ablation_guidance_scale}.
It can be seen that selecting a reasonable guidance intensity has an important role in learning the distinctive semantic features of malicious features. We ultimately choose $\beta=15$ as the optimal setting for guidance intensity.

\section{Artistic Style Erasure} \label{supp_art}

For evaluating the artistic style erasure ability, we follow the experiment settings from previous works~\cite{esd,uce,rece} using two benchmarks: Famous Artists~(Erase Van Gogh) and Modern Artists~(Erase Kelly Mckernan). Each dataset contains 20 prompts per artist style. We follow the previous works to measure the erasure ability of target artists~(Van Gogh and Kelly Mckernan) while measuring the preservation ability of unrelated styles.

\noindent \textbf{Evaluation benchmark.} We follow the setting from previous works~\cite{esd,uce,rece} to erase the style of artists ``\textit{Van Gogh}" and ``\textit{Kelly McKernan}," evaluating whether the methods can erase the target artistic style while retaining others. 

\begin{figure}[t]
    \centering
    \includegraphics[width=1\linewidth]{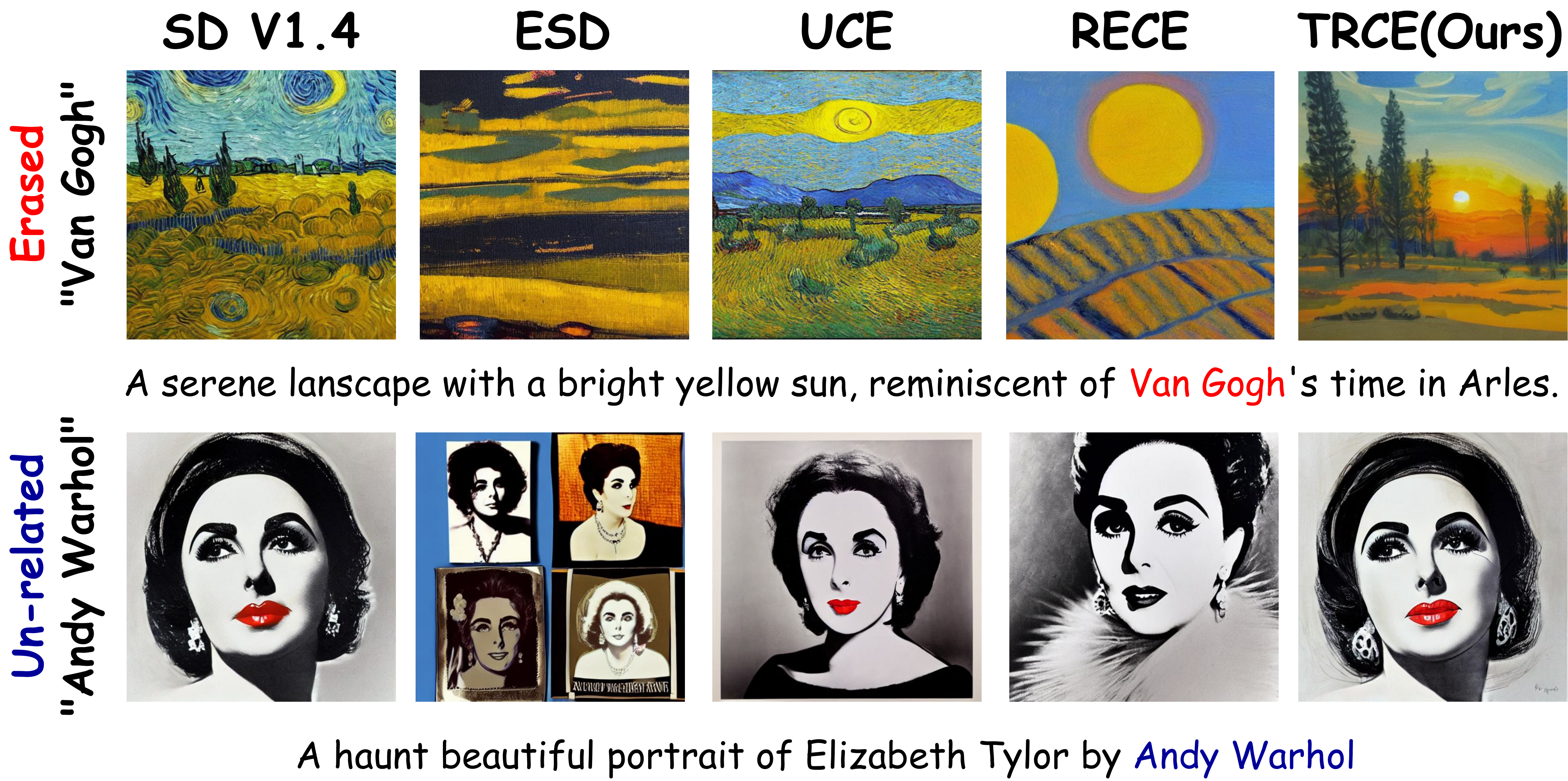}
    \caption{The visualization of artistic style erasure comparison. TRCE is able to effectively remove target styles while better preserving the details of the original image and prompt.
    }
    \label{fig:style}
\end{figure}

\begin{table}[t]
\centering
\small
\renewcommand\arraystretch{1.2}
\begin{tabular}{l|cc|cc}
\hline
\multirow{2}{*}{\textbf{Method}} & \multicolumn{2}{c|}{\textbf{Van Goah}} & \multicolumn{2}{c}{\textbf{Kelly Mckernan}} \\ \cline{2-5} 
                  & \textbf{Acc$_{e}$} $\downarrow$ & \textbf{Acc$_{u}$} $\uparrow$ & \textbf{Acc$_{e}$} $\downarrow$ & \textbf{Acc$_{u}$} $\uparrow$ \\ \hline
SD1.4~\cite{sd14}            & \textcolor{gray}{0.95}           & \textcolor{gray}{0.95}          & \textcolor{gray}{0.90}        & \textcolor{gray}{0.93}        \\
ESD~\cite{esd}                 & \textbf{0.15}        & 0.67        & \textbf{0.25}        & 0.70        \\
UCE~\cite{uce}               & 0.90        & \textbf{0.88}        & 0.75        & \textbf{0.85}        \\
RECE~\cite{rece}                & 0.35        & 0.83        & 0.30        & 0.80        \\
    
TRCE~(Ours) & \underline{0.20} & \underline{0.85}        & \textbf{0.25}  & \textbf{0.85}        \\ \hline
\end{tabular}
\caption{Comparison of artist style erasure task. The metric with SD V1.4 are reported for performance reference.}
\label{tab:style}
\end{table}

\noindent \textbf{GPT as style judger}: Given the subjective nature of artistic styles, followed by ~\cite{yoon2024safree}, we employ an advanced multi-modal large language model, GPT-4-o, as the judger to determine whether an image belongs to a specific artistic style. 

\noindent \textbf{Result analysis.} From the quantitative results illustrated in Table.~\ref{tab:style}, TRCE achieves favorable $Acc_e$ while effectively preserving un-related art styles~(as the evaluation performance of $Acc_u$).
In Fig.~\ref{fig:style}, we demonstrate the effect of erasing ``Van Gogh". We find that the advantage of TRCE in style erasure task is that it can erase the targeted style while preserving the original content and composition of the image, while better refer the prompt's instructions for generating the image content. 
This also indicates that TRCE better preserves the model's original ability to prevent unrelated content from being influenced.

\section{Celebrity Erasure} \label{supp_cele}

For evaluating the portrait protection ability of TRCE, we follow the prior work MACE~\cite{mace} to conduct experiments on erasing multiple celebrities. We select the most challenging benchmark of MACE that eliminates the concepts of 100 celebrities, which includes two subsets: 100 celebrities to be erased and 100 unrelated celebrities used to test the model's ability to retain knowledge.

\begin{figure}[t]
    \centering
    \includegraphics[width=1\linewidth]{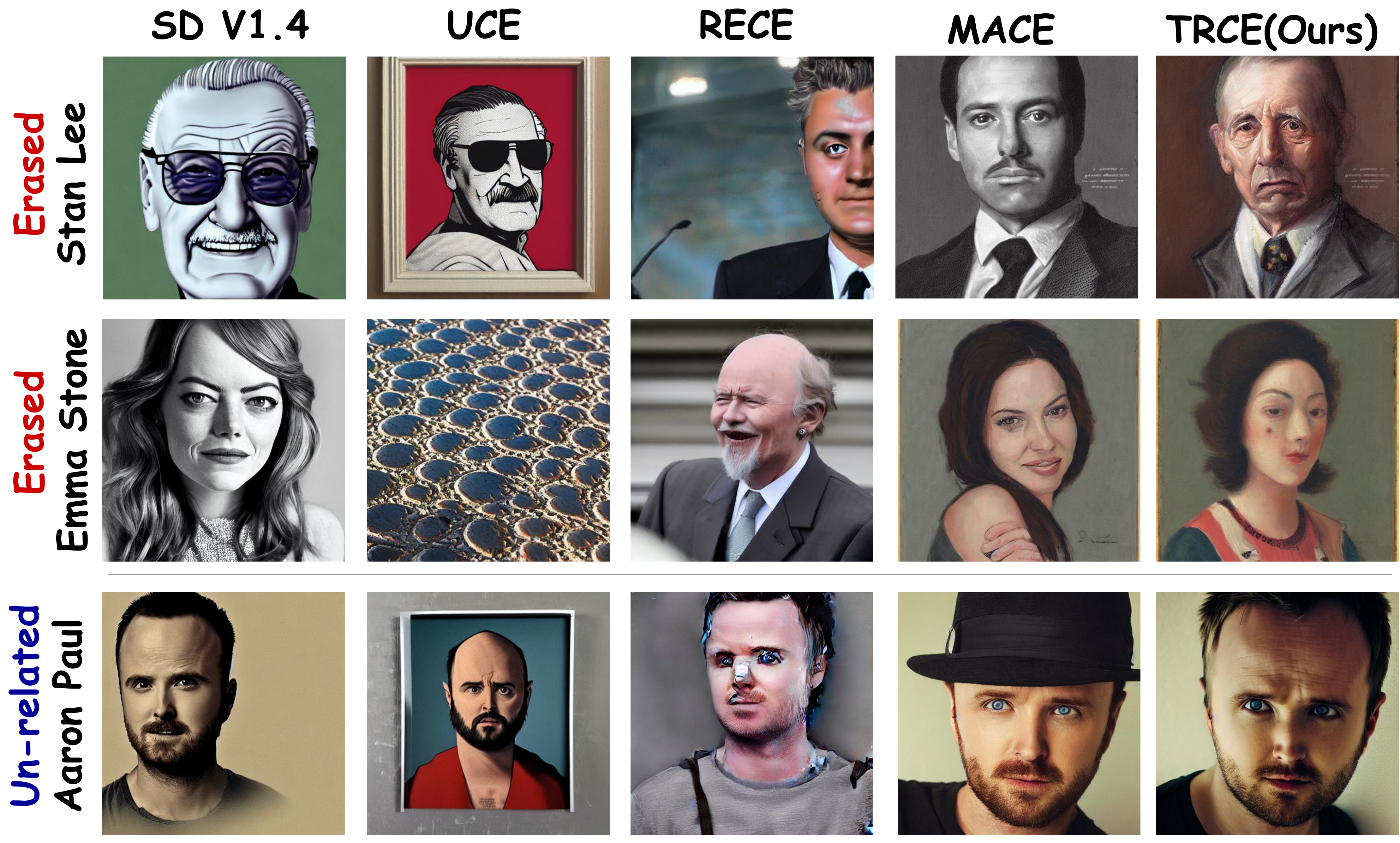}
    \caption{The visualization of celebrity erasure comparison. The prompts used to generate images are ``A portrait of \{\textit{the celebrity}\}".
    }
    \label{fig:cele}
\end{figure}

\begin{table}[t]
\centering
\renewcommand{\arraystretch}{1.1}
\begin{adjustbox}{width=\columnwidth}
\begin{tabular}{l|ccc|ccc}
\toprule
&Acc$_{e} \downarrow$ & Acc$_{r} \uparrow$ & $H_c \uparrow$  & $\text{FID}_{\mathrm{gen}}\downarrow$ & $\text{FID}_{\mathrm{real}}\downarrow$  &\text{CLIP-S} $\uparrow$\\
\midrule
UCE*~\cite{uce}     & 20.41 & 33.28  & 46.93 & 27.85 & \textbf{12.44} & 30.11 \\
RECE*~\cite{rece}   & 23.98  & 37.85 & 50.54 & 50.53 & 13.36 & 29.32 \\
MACE~\cite{mace}    & \textbf{3.52}  & 81.81 & 88.54 & \textbf{25.27} & 15.39 & 29.51 \\
\textbf{TRCE(Ours)} & 5.11& \textbf{85.32} & \textbf{89.85} & 25.29 & 12.79 & \textbf{30.48}\\
\bottomrule
\end{tabular}
\end{adjustbox}
\setlength{\abovecaptionskip}{0mm}
\caption{The evaluation result of erasing 100 celebrities on ~\cite{mace} dataset. The * tag indicates we reimplement these methods for comparison.}
\label{tab:celebrity}
\end{table}

\noindent \textbf{Metrics.} Following MACE, we use GIPHY Celebrity Detector~\cite{hasty2020giphy} whether the generated images reflects the target celebrities. We measure the erasing accuracy $\text{Acc}_e$, retaining accuracy $\text{Acc}_r$, and the harmonic mean $H_c$ of erasing and rataining, which is calculated as:
\begin{equation}
H_c = \frac{2}{(1 - \mathrm{Acc}_e)^{-1} + (\mathrm{Acc}_s)^{-1}}.
\end{equation}

\noindent \textbf{Implementations}. The implementation of celebrity erasure is basically similar to the multi-malicious concept erasure (discussed in Sec.\ref{append:implementation}). We generate 5 denoising trajectories for each celebrity for the second stage fine-tuning. In particular, considering that portrait generation shares similar visual patterns, we did not adopt the setting of only fine-tuning the visual layers in the second stage for the celebrity erasure. Instead, we only fine-tuned the text-related cross attention layers~(the key, value matrices) to avoid affecting the generation of irrelevant portraits.

\noindent \textbf{Result analysis.} The quantitative results are shown in Table.~\ref{tab:celebrity}, and Fig.~\ref{fig:cele} presents some visual comparison results. From the results, it can be seen that even though TRCE is not specifically designed for the erasure of massive portrait concepts, through the two-stage collaborative design, TRCE can still effectively erase concepts while preserving unrelated concepts without being affected.

\begin{figure}[t]
  \centering
  \setlength{\abovecaptionskip}{1mm}
  \includegraphics[width=1\linewidth]{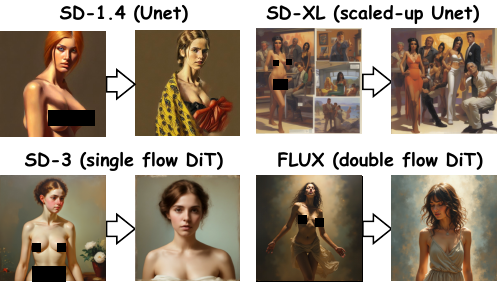}
   \caption{The proposed TRCE can transfer to various diffusion model architectures.}
   \label{fig:newmodel}
   \vspace{-4mm}
\end{figure}

\section{Compability with Newer Basemodel} \label{supp_basemodel}
As shown in Fig.~\ref{fig:newmodel}, we evaluate the compatibility of TRCE across different diffusion model architectures. The implementation solutions are as follows:

\noindent \textbf{Transfer to SD-XL model}. In the first-stage fine-tuning, since the UNet in SD-XL~\cite{sdxl} has more layers compared to SD1.4, we apply Textual Semantic Erasure only to the UNet encoder to reduce computational cost and minimize the impact on model capacity. In the second stage, we basically follow the settings from Sec.~\ref{append:implementation} and fine-tune the model using LoRA.

\noindent \textbf{Transfer to DiT-based model}. For DiT-based models such as SD-3~\cite{sd3} and FLUX-dev~\cite{blackforestlabs2024flux1dev}, which do not have separate cross-attention layers for handling textual information, we follow the latest UCE implementation and apply first-stage erasure on the ``\textit{context\_embedder}" layer. In the second stage, we also use LoRA for model fine-tuning.

\section{More Visualization Results} \label{sup_vis}
\begin{figure*}[t]
    \centering
    \includegraphics[width=1\textwidth]{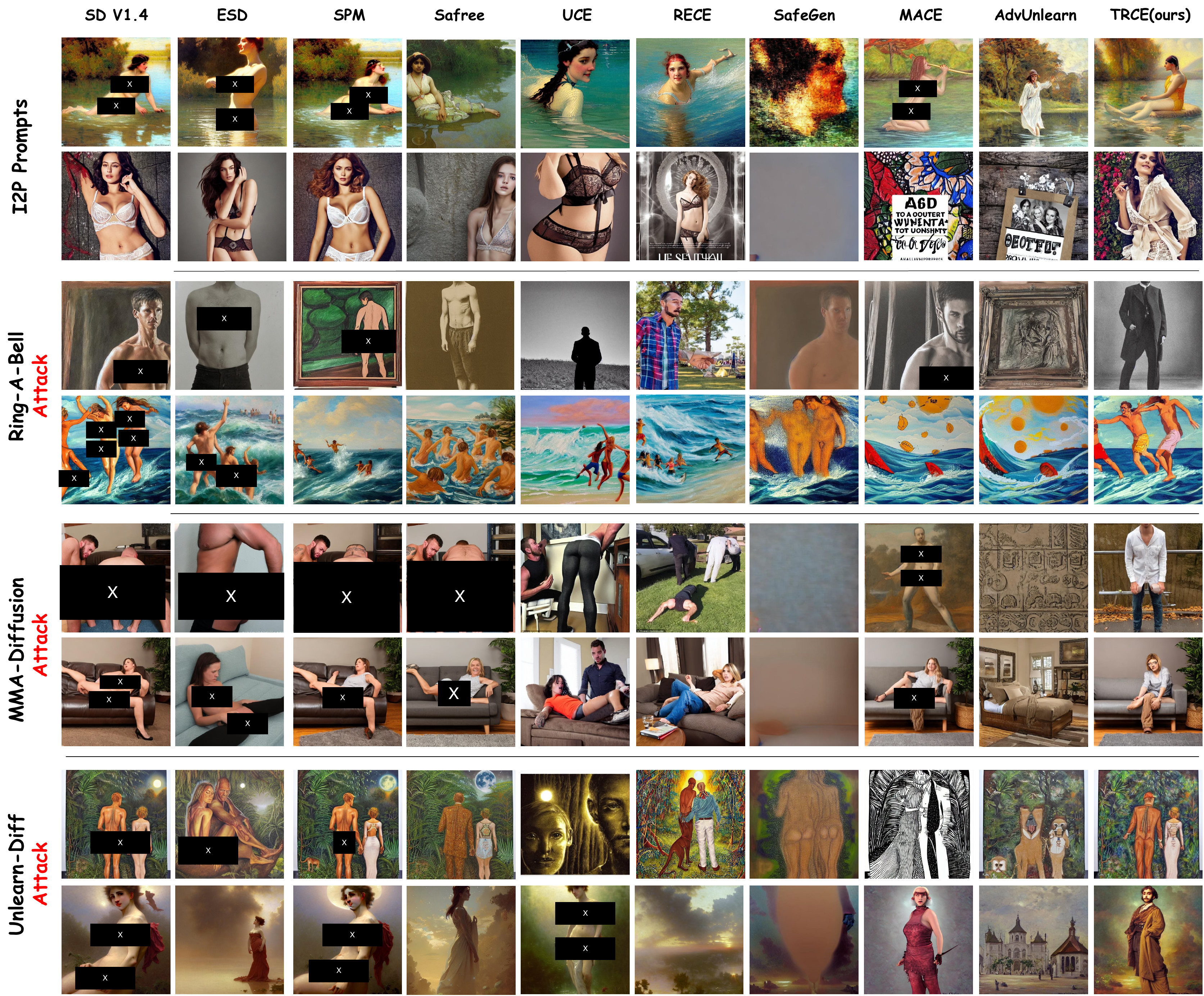}
    \captionof{figure}{The visualization of the erasure ability of current methods on I2P~\cite{sld}, Ring-A-Bell~\cite{ring-a-bell}, MMA-Diffusion~\cite{mma} and Unlearn-Diff~\cite{unlearndiff} datasets. TRCE achieves a reliable ``sexual" concept erasing while maintaining the overall visual context of generated images.}
    \label{compare_more}
\end{figure*}

\begin{figure*}[t]
    \centering
    \includegraphics[width=1\textwidth]{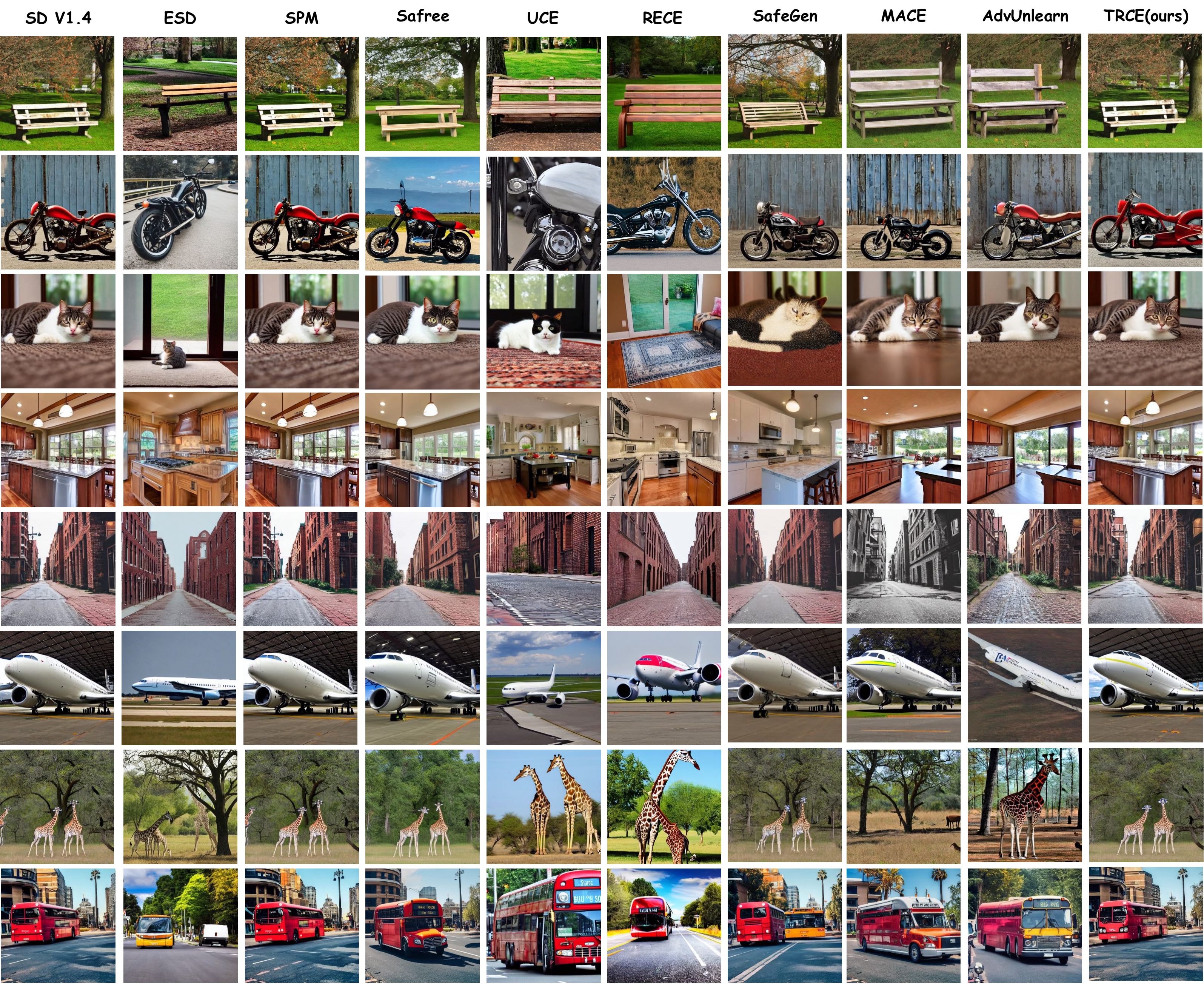}
    \captionof{figure}{The visualization of knowledge preservation ability to generate general images~\cite{coco}. TRCE better preserves the generation of general images and exhibits a strong knowledge preservation ability.}
    \label{more_knowledge}
\end{figure*}

\begin{figure*}[t]
    \centering
    \includegraphics[width=1\textwidth]{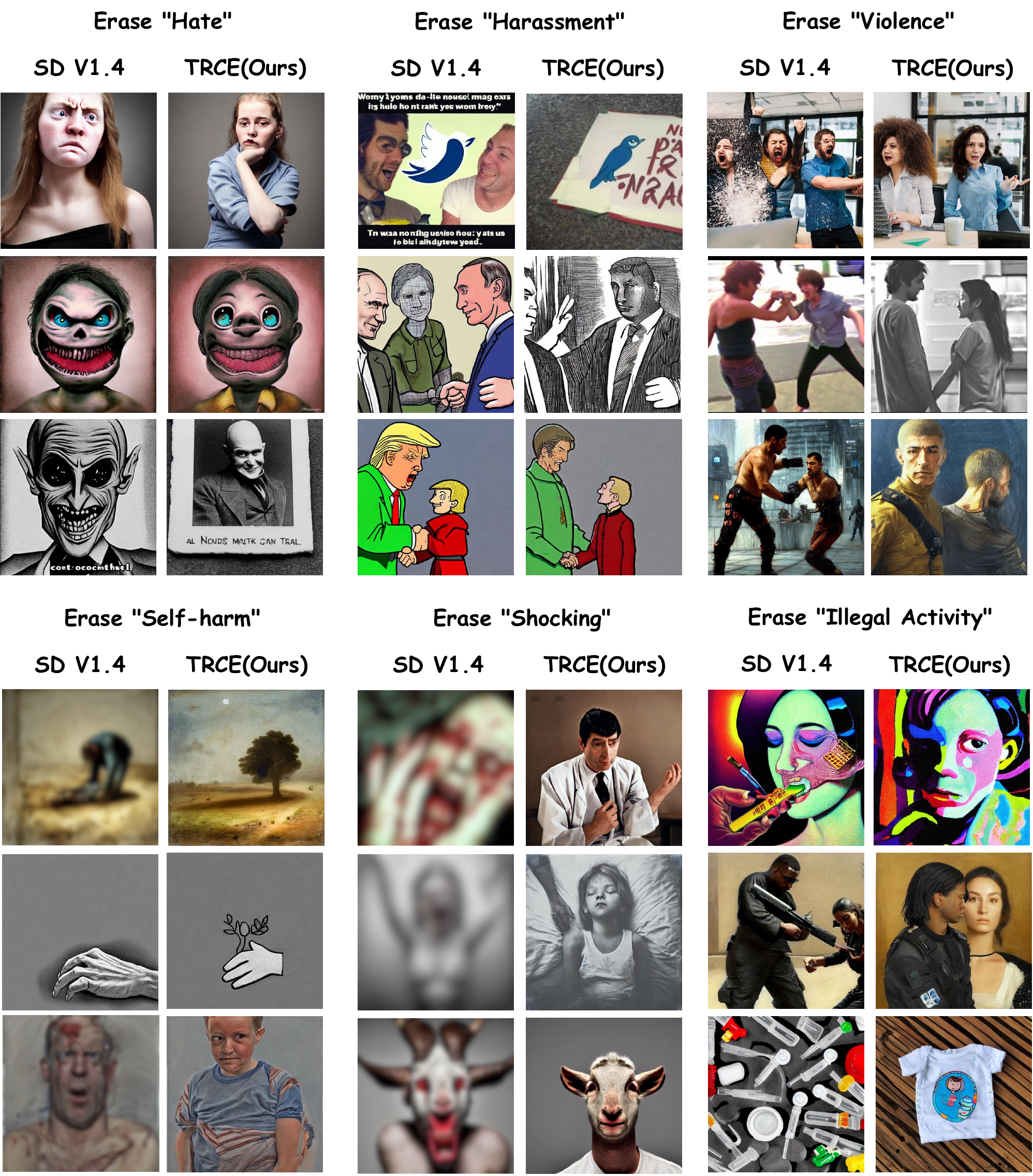}
    \captionof{figure}{The visualization of the erasure ability of TRCE on simultaneously erasing multiple malicious concepts in the I2P dataset.}
    \label{more_unsafe}
\end{figure*}

In this part, we showcase more visualization results. In Fig.~\ref{compare_more}, we display erasure comparison through different benchmarks~\cite{sld,ring-a-bell,mma,unlearndiff}. Fig.~\ref{more_unsafe} showcases some additional visualization results on multiple malicious concept erasure. And finally, Fig.~\ref{more_knowledge} provides visualization results of knowledge preservation comparison.

